\documentclass{article}

 \usepackage[preprint]{neurips_2025}

\usepackage[utf8]{inputenc} %
\usepackage[T1]{fontenc}    %
\usepackage{hyperref}       %
\usepackage{url}            %
\usepackage{booktabs}       %
\usepackage{amsfonts}       %
\usepackage{nicefrac}       %
\usepackage{microtype}      %
\usepackage{xcolor}         %

\usepackage{wrapfig}
\usepackage{enumitem}
\usepackage{comment}
\usepackage{graphicx}
\usepackage{multirow} %
\usepackage{xspace}
\usepackage{pifont}
\usepackage{caption}
\usepackage{setspace}
\usepackage[most]{tcolorbox}

\usepackage{listings}
\usepackage[most]{tcolorbox}

\usepackage{cleveref}

\usepackage{subcaption}
\usepackage{dcolumn}

\usepackage{placeins}

\lstdefinestyle{neuprompt}{
  basicstyle=\ttfamily\small,
  columns=fullflexible, keepspaces=false,
  tabsize=1, showstringspaces=false,
  breaklines=true, breakautoindent=false, breakindent=0pt,
  prebreak=\mbox{}, postbreak=\mbox{},
  aboveskip=0pt, belowskip=0pt   %
}

\newtcblisting{promptblock}{
  listing only, breakable, enhanced,
  colback=black!2, colframe=black!20,
  boxsep=0pt, left=6pt, right=6pt, top=6pt, bottom=6pt, arc=2pt,
  before skip=4pt plus 1pt minus 1pt,   %
  after  skip=4pt plus 1pt minus 1pt,
  listing engine=listings,
  listing options={style=neuprompt}
}

\title{Remote Labor Index:\\Measuring AI Automation of Remote Work}

\newcommand\blfootnote[1]{%
  \begingroup
  \renewcommand\thefootnote{}%
  \footnotetext{#1}%
  \addtocounter{footnote}{-1}%
  \endgroup
}

\author{
    \begin{minipage}{\textwidth} 
    \centering
    \setstretch{1.1}
    \hspace{-1em}Mantas Mazeika$^{*1}$,\; Alice Gatti$^{*1}$,\; Cristina Menghini$^{*\dagger}$, \\
    Udari Madhushani Sehwag$^{*2}$,\; Shivam Singhal$^{*\dagger}$,\; Yury Orlovskiy$^{*1}$\\[0.9em]
    \hspace{-1em}Steven Basart$^1$,\; Manasi Sharma$^2$,\; Denis Peskoff$^2$,\; Elaine Lau$^2$,\; Jaehyuk Lim$^1$, \\
    Lachlan Carroll$^1$,\; Alice Blair$^1$,\; Vinaya Sivakumar$^1$,\; Sumana Basu$^2$,\; Brad Kenstler$^2$, \\
    Yuntao Ma$^\dagger$,\; Julian Michael$^\dagger$,\; Xiaoke Li$^1$,\; Oliver Ingebretsen$^1$,\; Aditya Mehta$^1$, \\
    Jean Mottola$^1$,\; John Teichmann$^\ddagger$,\; Kevin Yu$^\ddagger$,\; Zaina Shaik$^\ddagger$,\; Adam Khoja$^1$, \\
    Richard Ren$^1$,\; Jason Hausenloy$^1$,\; Long Phan$^1$,\; Ye Htet$^2$,\; Ankit Aich$^2$, \\
    Tahseen Rabbani$^2$,\; Vivswan Shah$^\dagger$,\; Andriy Novykov$^1$,\; Felix Binder$^\dagger$ \\[0.9em]
    \hspace{-1em}Kirill Chugunov$^2$,\; Luis Ramirez$^2$,\; Matias Geralnik$^2$,\; Hernán Mesura$^2$, \\
    Dean Lee$^\dagger$,\; Ed-Yeremai Hernandez Cardona$^2$,\; Annette Diamond$^\dagger$ \\[0.9em]
    \hspace{-1em}Summer Yue$^{**\dagger}$,\; Alexandr Wang$^{**\dagger}$,\\ Bing Liu$^{**2}$,\; Ernesto Hernandez$^{**2}$,\; Dan Hendrycks$^{**1}$ \\[1.5em]
    \hspace{-2.7em}{\normalfont\mdseries$^1$Center for AI Safety \quad $^2$Scale AI}
    \end{minipage}
}

\begin{document}

\maketitle

\blfootnote{$^*$Equal contribution \quad $^{**}$Senior authors \quad $^\dagger$Work done while at Scale AI \quad $^\ddagger$Work done while at CAIS}

\vspace{-10pt}
\begin{abstract}
AIs have made rapid progress on research-oriented benchmarks of knowledge and reasoning, but it remains unclear how these gains translate into economic value and automation. To measure this, we introduce the Remote Labor Index (RLI), a broadly multi-sector benchmark comprising real-world, economically valuable projects designed to evaluate end-to-end agent performance in practical settings. AI agents perform near the floor on RLI, with the highest-performing agent achieving an automation rate of $2.5\%$. These results help ground discussions of AI automation in empirical evidence, setting a common basis for tracking AI impacts and enabling stakeholders to proactively navigate AI-driven labor automation.

\end{abstract}

\section{Introduction}

The potential for AI to automate human labor is a subject of profound societal interest and concern. As AI capabilities advance, understanding their impact on the workforce becomes increasingly urgent. However, we lack standardized, empirical methods for monitoring the trajectory of AI automation. Without reliable metrics grounded in real-world economic activity, stakeholders may struggle to build consensus and proactively navigate AI-driven labor automation.

While AI systems have demonstrated rapid progress on a variety of benchmarks, it remains unclear how these gains translate into the capacity to perform economically valuable work. Many existing AI agent benchmarks measure performance on specialized skills such as software engineering \citep{jimenez2023swe,miserendino2025swe,rein2025hcast} and basic computer use \citep{zhou2023webarena,deng2023mind2web,koh2024visualwebarena,mialon2023gaia,xie2024osworld}, while some focus on simple tasks shared across several professions \citep{patwardhan2025gdpval}. These provide valuable signals of capabilities in isolation, yet they often do not capture the vast diversity and complexity inherent in the broader landscape of remote work. Consequently, performance on these benchmarks offers limited insight into the trajectory of human labor automation.

\begin{figure}
  \centering
  \vspace{-30pt}
  \makebox[\linewidth][c]{ \includegraphics[width=1.1\linewidth]{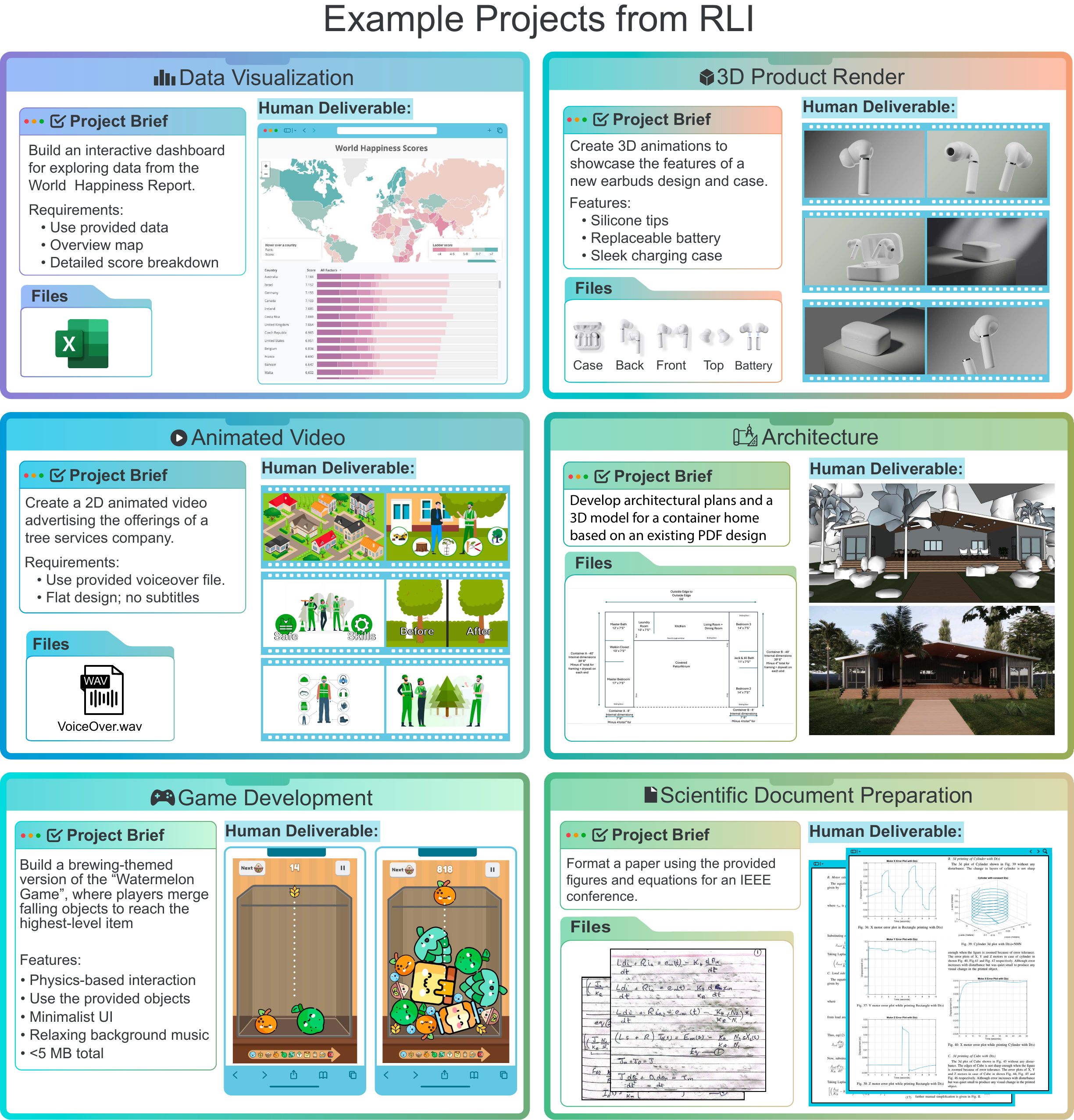}
  }
  \caption{The Remote Labor Index (RLI) represents a broad range of projects from across the remote labor economy, including game development, product design, architecture, and data analysis. All projects represent real work that was performed by human professionals.}\label{fig:splash}
  \vspace{-15pt}
\end{figure}

We introduce the Remote Labor Index (RLI) to provide the first standardized, empirical measurement of AI's capability to automate remote work. RLI is designed to evaluate AI agents on their ability to complete real-world, economically valuable work, spanning the large share of the economy that consists of computer-based work. RLI is composed of entire projects sourced directly from online freelance platforms, reflecting the diverse demands of the remote labor market. These projects exhibit significantly higher complexity than tasks found in existing agent benchmarks. Crucially, by sourcing the majority of projects from freelancing platforms, RLI is grounded in actual economic transactions, encompassing the original work brief and the gold-standard deliverable produced by a human freelancer. This structure allows for a direct assessment of whether AI agents can produce economically valuable work.

We evaluate several frontier AI agent frameworks on RLI, utilizing a rigorous manual evaluation process to compare AI outputs against the human gold standard. The results indicate that performance on the benchmark is currently near the floor. The best-performing current AI agents achieve an automation rate of $2.5\%$, failing to complete most projects at a level that would be accepted as commissioned work in a realistic freelancing environment. This demonstrates that despite rapid progress on knowledge and reasoning benchmarks, contemporary AI systems are far from capable of autonomously performing the diverse demands of remote labor. To detect more granular shifts in performance, we employ an Elo-based pairwise comparison system. While all models fall well short of the aggregate human baseline, we observe that models are steadily approaching higher automation rates across projects.

By introducing RLI, we aim to ground discussions of AI automation in empirical evidence and provide a common basis for understanding AI automation capabilities on economically valuable projects. We hope this provides an empirical foundation for researchers, policymakers, and the public to navigate the onset of AI automation of remote labor.

\section{Related Work}

\begin{figure}[t]
\vspace{-10pt}
\centering
\includegraphics[width=0.9\linewidth]{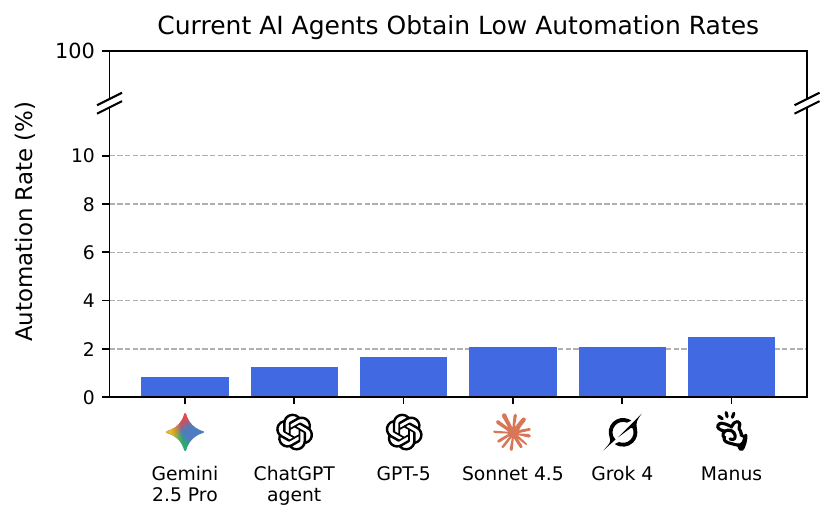}
\caption{All AI agents tested automate at most $2.5\%$ of tasks on RLI, showing that most economically valuable remote work currently remains far beyond their capabilities.}
\label{fig:main-bar-chart}
\end{figure}

\paragraph{Evaluating AI agents.}

The potential impact of AI automation on the global economy and labor markets has been the subject of significant economic analysis \citep{brynjolfsson2025canaries,acemoglu2025simple}. Complementing this macroeconomic perspective, the machine learning community has increasingly focused on empirically measuring AI's capacity to perform economically valuable work. The scope of benchmarks evaluating AI systems on valuable work has expanded considerably over time. Efforts have broadened from evaluating closed-ended academic knowledge \citep{hle,glazer2024frontiermath,rein2024gpqa,hendrycks2020measuring} to include agentic tasks that require interaction with dynamic environments. This shift encompasses autonomous computer use \citep{xie2024osworld,mialon2023gaia,liu2023agentbench}, web browsing \citep{zhou2023webarena,deng2023mind2web,koh2024visualwebarena}, and realistic API calls \citep{yao2024tau}.

\paragraph{Benchmarking real-world value.}

Knowledge benchmarks at the limits of human skill are becoming saturated, and current agent benchmarks often rely on simplified environments, representing only a small fraction of the remote work economy. There have been a number of domain-specific benchmarks measuring specific kinds of work, including software engineering \citep{jimenez2023swe,miserendino2025swe,rein2025hcast}, ML engineering \citep{chan2024mle,wijk2024rebench,edwards2025rexbench,starace2025paperbench}, and others \citep{penrose-accounting,vidgen2025apex}. Most similar to our work, Patwardhan et al. \citep{patwardhan2025gdpval} show AI models are near human parity on specific kinds of tasks shared across a wide range of professions, such as writing, web search, and administrative tasks. This indicates that current AIs have significant potential for augmentation but does not enable measuring the capacity for end-to-end project automation.

In contrast to prior benchmarks, RLI measures the automation ability of AI agents on end-to-end projects sourced from real-world work in remote labor markets, thereby grounding the evaluation in actual economic transactions. Hendrycks et al.~\citep{hendrycks2025agidefinition} measure general human-level cognitive ability representing well-educated individuals, whereas RLI targets automation capacity relative to the remote work economy, which is an aggregate of diverse human specializations and skills.

\section{Remote Labor Index}

We introduce the Remote Labor Index (RLI), a new benchmark composed of end-to-end remote freelance projects for evaluating AI agents on practical, economically valuable work. Our data is sourced directly from professionals on freelance platforms, grounding the benchmark in economic value and capturing the diversity and complexity of real remote labor markets. The final dataset comprises $240$ projects.

\subsection{Dataset Description} \label{sec:dataset_description}

Here, we describe the contents of RLI projects and high-level statistics of the data. More details on these topics are available in Appendix \ref{app:dataset_details}.

\paragraph{Project composition.}
Each project in RLI consists of three components:
\begin{itemize}
    \item \textbf{Brief}: A text document describing the work to be done
    \item \textbf{Input files}: A directory containing files needed to complete the project
    \item \textbf{Human deliverable}: A gold-standard deliverable that successfully completes the project, produced by a professional
\end{itemize}

These components are visualized for a sample of projects in Figure \ref{fig:splash}. For each project, the brief and input files are provided by the professional who produced the human deliverable. This ensures the brief and input files contain sufficient information to complete the project. For each project, we also record the time and cost to produce the gold-standard human deliverable, as reported by the professional who carried out the work.

\begin{wrapfigure}{r}{0.5\textwidth}
  \vspace{-20pt}
  \centering
  \includegraphics[width=0.48\textwidth]{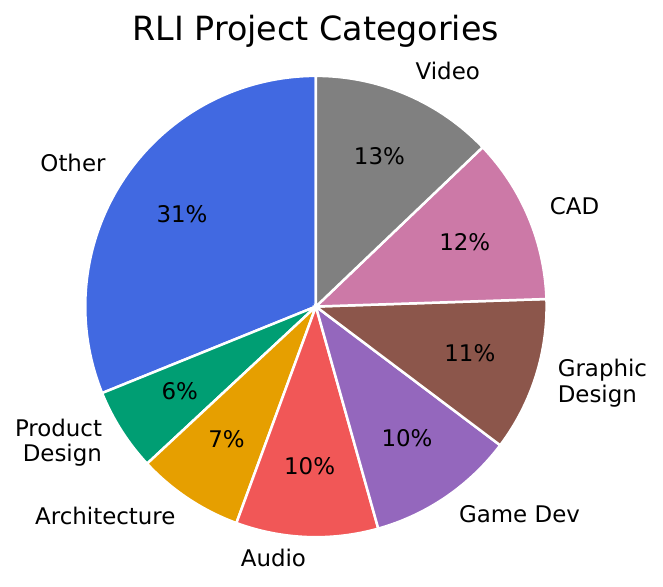}
  \caption{RLI captures a wide array of project types, spanning $23$ categories of work from the Upwork taxonomy. Here, we show the top seven categories.}
  \label{fig:pie-chart}
  \vspace{-10pt}
\end{wrapfigure}

\paragraph{Coverage of types of work.}
RLI is diverse along two axes central to real knowledge work: (i) the range of jobs represented (measured by the Upwork taxonomy) and (ii) the file formats of the artifacts required to complete them. 
The Upwork taxonomy is well-suited for end-to-end remote freelance labor. In preliminary analysis, we found that the O*NET taxonomy, while valuable for long-term occupations, was less tailored to the remote labor markets represented in RLI (see Appendix \ref{app:onet_limitations}). Following the collection and review process detailed in Section \ref{sec:dataset_collection}, our final dataset covers $23$ categories of work out of Upwork's $64$. These categories are reported in Appendix \ref{app:upwork_domains}. In addition, the input files and deliverables in RLI span a wide variety of file types (\Cref{fig:filetype_comparison}), substantially more than previous comparable benchmarks.

A useful lens on project composition is the distinction between software/research/writing tasks and the wider landscape of remote labor. Prior agent benchmarks tend to emphasize the former, where today’s models already perform relatively well. As \Cref{fig:upwork_comparison} shows, however, real freelance remote labor is far less concentrated in these activities. RLI is designed for this broader reality: it includes substantial coverage of design, operations, marketing, administration, data/BI, audio–video production, and other categories, sampling across task complexity and deliverable types to reflect end-to-end freelance remote labor.

\paragraph{Difficulty and economic value.}
Finally, we report the effort required to produce the gold-standard human deliverables. As shown in \Cref{fig:upwork_comparison}, the completion time for RLI projects exceeds previous benchmarks by more than $2\times$, with a mean of $28.9$ hours and median of $11.5$ hours. This matches the completion time of a random sample of jobs on Upwork, demonstrating how RLI comes closer than previous benchmarks to capturing the true complexity of remote labor markets. The average cost of projects in RLI is $\$632.6$ with a median of $\$200$. Taken together, these properties yield a benchmark that is challenging and, in aggregate, more representative of contemporary remote freelance work than previous benchmarks. For more details on the dataset cost and time, see Appendix \ref{app:time_cost_distribution}

\subsection{Dataset Collection} \label{sec:dataset_collection}

\begin{figure}[t]
  \vspace{-15pt}
  \centering
  \includegraphics[width=1.0\linewidth]{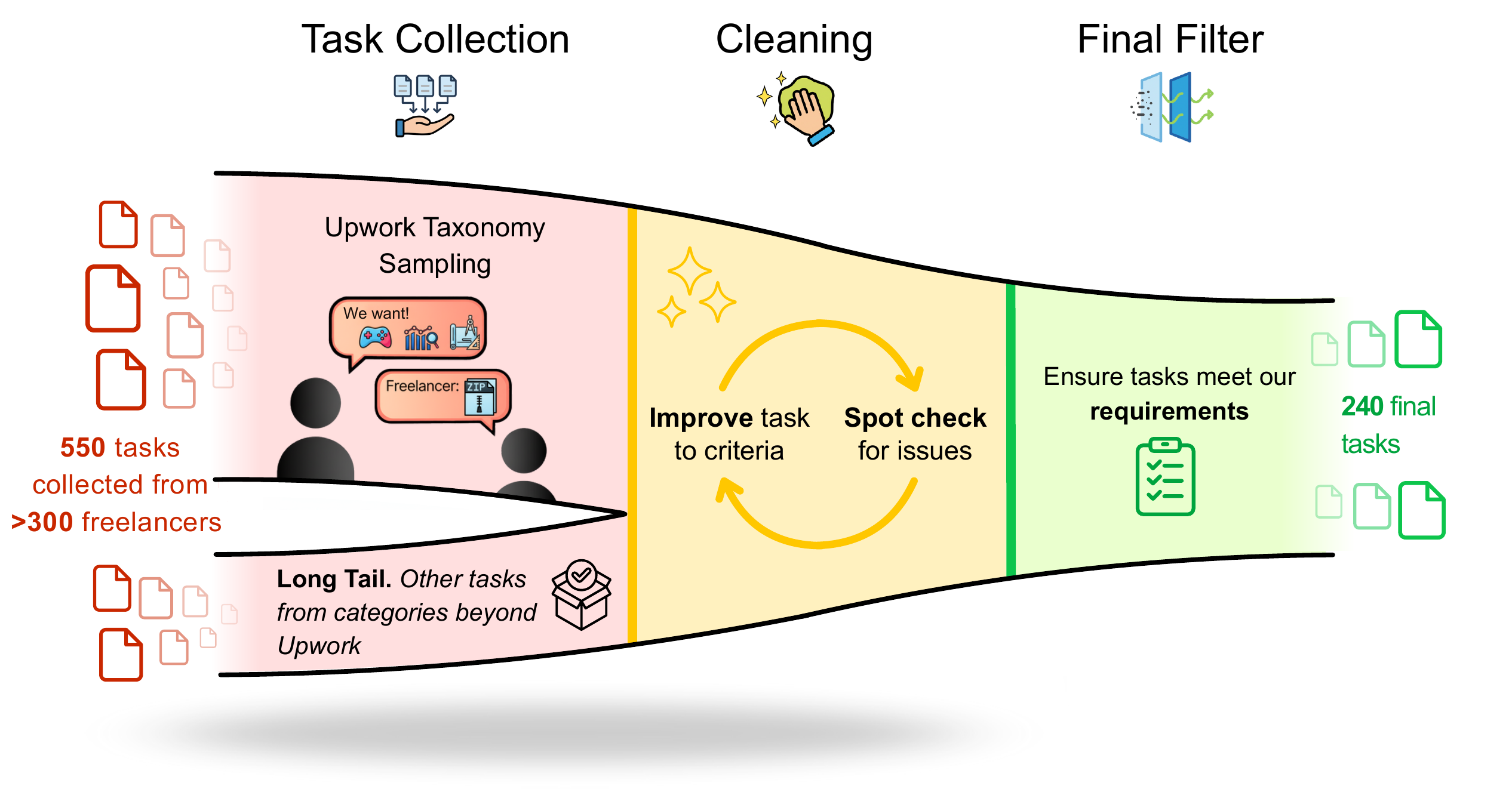}
  \caption{RLI projects were extensively filtered and cleaned to ensure quality. Projects were sourced primarily from the remote labor market and secondarily from deliverables representing uncommon and emerging types of remote work work. (For details, see Appendix~\ref{app:dataset_details}.)}
  \label{fig:collection_pipeline}
\end{figure}

Here, we describe how the data were collected, the expertise of the contributors, and the cleaning process. The full pipeline is visualized in Figure \ref{fig:collection_pipeline}.

\begin{figure}[t]
\vspace{-10pt}
\centering
\includegraphics[width=1.0\linewidth]{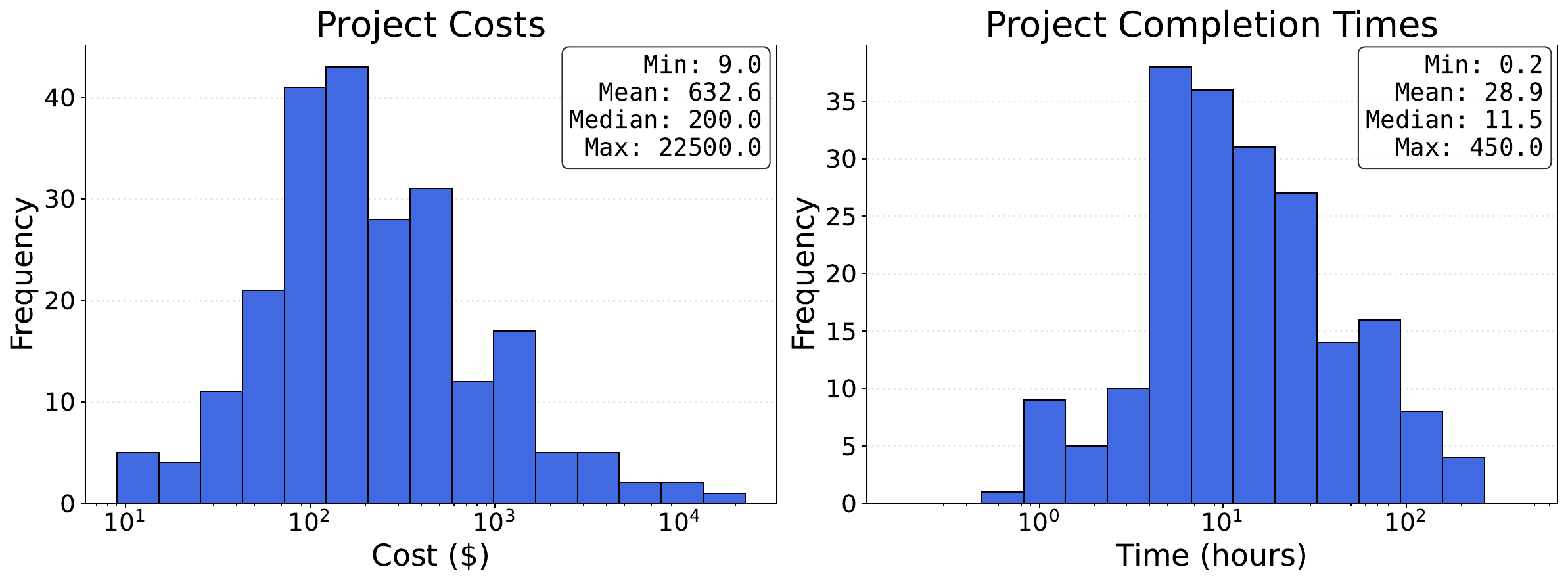}
\caption{RLI spans a broad range of difficulty, with project costs reaching over $\$10,\!000$ and completion times for human professionals reaching over $100$ hours. All project costs and completion times come directly from human professionals who completed the projects. In total, the projects in RLI represent over $6,\!000$ hours of real work valued at over $\$140,\!000$.}
\label{fig:completion-time-cost}
\end{figure}

\paragraph{Sourcing strategy and scope.}
Our collection methodology is bottom-up, engaging directly with human professionals who were willing and authorized to provide their past work samples for our research. This approach ensures that our projects reflect genuine market demands and complexities.

We defined the scope of collection using the Upwork taxonomy. Starting from the full list of $64$ categories, we filtered out categories that did not meet predefined criteria necessary for a standardized benchmark. For example, we excluded work requiring physical labor (e.g., local photography), work that requires waiting to evaluate (e.g., SEO), or work that cannot be easily evaluated in a web-based evaluation platform (e.g., back-end development). For the full set of exclusion criteria, see \Cref{app:task_filtering}. This filtering resulted in $43$ eligible categories.

We sourced projects in two stages:

\begin{enumerate}
    \item \textbf{Freelance Platform Sourcing}: We submitted a job post for each category within the $43$ eligible categories (e.g., 3D animation, Mechanical Engineering, Presentation Design; the full list is in Appendix \ref{app:data_collection}). Hired freelancers provided samples of their prior work, yielding a diverse pool of projects. In total, this yielded $207$ projects.
    
    \item \textbf{Long-Tail Sourcing}: Digital labor marketplaces contain a substantial long tail of work. To sample from this long tail, we hired freelancers to provide work samples from additional categories not in the Upwork taxonomy and commissioned custom work. In total, this yielded $7$ projects. We also expanded beyond Upwork, identifying high-quality examples of digital work available online. For these examples, we contacted the authors to request permission to use their work in our study and to ascertain the time taken and the monetary value of their labor on the project. We only include projects where authors gave permission and provided this timing and pricing information, yielding an additional $33$ projects.
\end{enumerate}

\paragraph{Recruitment and expertise.}
We recruited $358$ freelancers with verified Upwork accounts and specialization in the target categories. These professionals demonstrated significant experience: on average, they had $2,\!341$ hours worked, $89$ prior jobs, and $\$23,\!364$ in total earnings on Upwork. From these freelancers, we collected $550$ initial projects. Freelancers were paid between $\$15$ and $\$200$ per project (average $\$41$) to sell us existing work samples.

\begin{figure}[t]
  \centering
  \includegraphics[width=\linewidth]{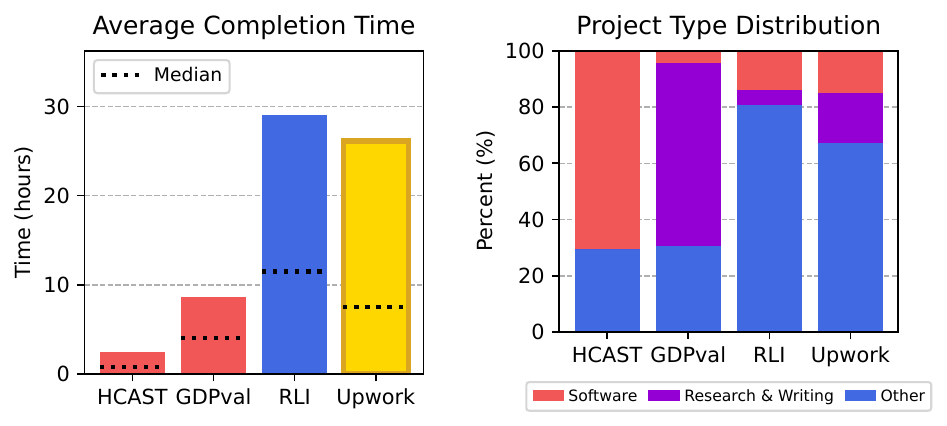}
  \caption{RLI is far closer to the complexity and diversity of real freelance labor than previous comparable benchmarks. Left: The average completion time for humans on RLI projects matches the true Upwork distribution. Right: Previous benchmarks primarily focus on tasks involving software engineering or web-based research and writing, but real remote labor markets have far more diversity.}
  \label{fig:upwork_comparison}
\end{figure}

\paragraph{Review and cleaning.}
To ensure each project is a self-contained, reproducible benchmark instance, we conducted multiple rounds of review, cleaning and standardization (Figure \ref{fig:collection_pipeline}). In each review, we  carefully evaluated the brief, input materials, and deliverables for suitability. We excluded project types that failed to meet our criteria (see Appendix \ref{app:task_filtering}). Examples include projects requiring human interaction or those producing deliverables in proprietary formats that could not be readily rendered for evaluation (see Section \ref{sec:evaluation}). When needed, we followed up with freelancers for clarifications or missing materials. We then normalized all accepted projects to a common schema and, in a final pass, removed additional projects that were ultimately unsuitable. Although this rigorous multi-step filtering process slightly shifted the final project distribution, the resulting benchmark remains a highly representative and challenging sample of remote knowledge work (see \Cref{fig:upwork_comparison}).

\paragraph{Data privacy and release.}
The final RLI dataset contains $240$ projects. To protect PII and prevent benchmark contamination, we maintain a private test set of $230$ projects used for quantitative evaluation. We release a public set of $10$ projects along with the open-sourced code for the evaluation platform to enable qualitative evaluation. None of the project descriptions in RLI are searchable. For the long-tail data, some human deliverables exist online, but not in a form that can be downloaded and presented as the full deliverable. To further protect against contamination in these cases, we include a blocklist of domains.

\subsection{Metrics}

We use the following metrics to measure performance on RLI for a given AI agent:
\begin{itemize}
\item \textbf{Automation rate}: The percentage of projects for which the AI deliverable is judged by human evaluators to complete the project at least as well as the human deliverable. This measures the absolute success rate of the AI agent across RLI projects.
\item \textbf{Elo}: A score capturing the relative performance of different AI agents. For each project, a deliverable from two different AIs is presented to human evaluators, who judge which deliverable is closer to completing the project successfully. If both agents successfully complete the project, then their deliverables are compared on overall quality. A difference of $400$ corresponds to $10\!:\!1$ odds of winning.
\item \textbf{Dollars earned}: The combined dollar value of the projects successfully completed by the AI agent, using the cost of the human deliverable $\texttt{cost}(H)$ as the dollar value for each project. The profit earned from completing all projects would be $\$143,991$.
\item \textbf{Autoflation}: The percentage decrease in the cost of completing the fixed RLI project bundle when using the cheapest-possible method to complete each project (human deliverable or an AI deliverable). We compute this as $1 - \frac{\sum \min\!\big(\texttt{cost}(H),\, \min_j \texttt{cost}(AI_j)\big)}{\sum \texttt{cost}(H)}$, where $\texttt{cost}(H)$ is the cost of the human deliverable and $\texttt{cost}(AI_j)$ is the cost of an evaluated AI agent solving the project. In cases where the AI deliverable does not complete the project, we set $\texttt{cost}(AI_j) = \infty$. This metric is discussed further in Appendix \ref{app:autoflation}.

\end{itemize}

The automation rate and Elo metrics are fully compatible, in that automation rate equals the probability of a win or tie against the human baseline under the same standards as the Elo evaluation. This allows computing an Elo score for the human baseline. We canonicalize scores so that the human baseline Elo is fixed at $1,\!000$.

\subsection{Evaluation} \label{sec:evaluation}

The deliverables in RLI are complex and span a wide range of formats. Evaluating these deliverables is itself a demanding task, often requiring on-the-job learning, complex computer use, and lengthy multimodal analysis. As this level of assessment is currently beyond the capabilities of automated evaluation systems, we rely on rigorous manual evaluation. This section details the process for generating AI deliverables, the platform used for evaluation, and the methodologies for assessing both the automation rate and Elo scores.

\paragraph{Deliverable generation.}
To generate deliverables, agents are provided with the project brief and input files. We do not mandate a specific execution environment or agent architecture. However, to ensure that the resulting artifacts can be properly assessed, agents receive an evaluation compatibility prompt before beginning the project. This prompt details the capabilities of our evaluation platform and provides a comprehensive, readable list of supported file formats, guiding the agent to produce outputs that are renderable and reviewable. The specific agents used for our pre-release evaluation are described in Appendix \ref{app:agent-environments}.

\paragraph{Evaluation platform.}
To standardize the review process and manage the diversity of deliverable formats, we developed a specialized web-based evaluation platform (an example is shown in Appendix \ref{app:evaluation_platform_details}). This platform allows evaluators to efficiently explore unstructured deliverable directories and natively render dozens of different file formats, facilitating a consistent evaluation experience across varied projects. The code for the evaluation platform is open-sourced.

\begin{figure}[t]
  \vspace{-10pt}
  \centering
  \includegraphics[width=1\linewidth]{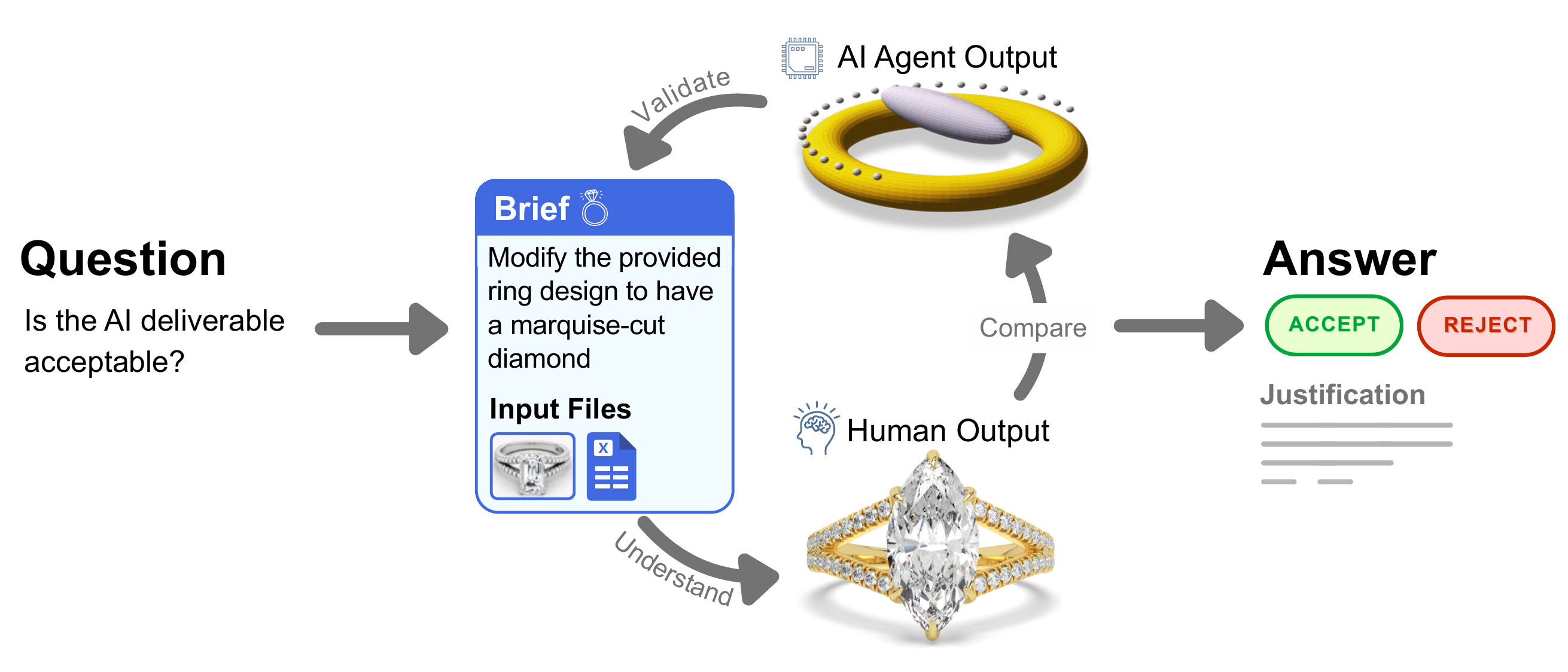}
  \caption{Evaluation Pipeline: For each RLI project, AI deliverables are rigorously checked against human gold-standard deliverables and the requirements in the project brief for flaws and to determine whether the AI deliverable would be accepted as work product in a realistic freelance setting. Evaluating AI deliverables is itself a highly agentic task, so automating evaluation with LLMs is not currently feasible. Thus, all evaluations are performed manually by trained workers and subject experts. Inter-annotator agreement is above $94\%$.}
  \label{fig:evaluation_pipeline}
\end{figure}

\paragraph{Automation rate evaluation.}
Our evaluation methodology centers on determining whether an AI deliverable completes the project at least as well as the human gold standard—specifically, whether the deliverable would be accepted by a reasonable client as the commissioned work.

In preliminary evaluations, we found granular per-project rubrics were often insufficient for capturing project completion. Particularly for projects with hard-to-specify aspects (e.g., design), a deliverable might technically satisfy rubric elements yet fail professional standards. Consequently, we employ a holistic evaluation approach (visualized in Figure \ref{fig:evaluation_pipeline}), drawing from practices for reviewing complex artifacts like papers or grants. Evaluators digest the project context (brief, input files, human deliverable) and compare the human and AI deliverables, examining specific files until confident in their assessment. Given a fixed time per project, they assess the AI deliverable (the alternative) relative to the human deliverable (the reference) using the following $3$-point scale, with a written justification:
\begin{enumerate}
    \item The alternative deliverable does not satisfy the brief as well as the reference deliverable or is of significantly lower quality, such that it would not be accepted by a reasonable client as the commissioned work.
    \item The alternative deliverable satisfies the brief as well as the reference deliverable and would be accepted by a reasonable client as the commissioned work.
    \item Same as $2$, and the alternative deliverable exceeds the reference deliverable in overall quality.
\end{enumerate}
The automation rate is calculated based on the percentage of projects receiving an annotation of $2$ or $3$. This holistic approach allows for targeted analysis, enabling evaluators to ``zoom into the deliverable'' and quickly identify major issues without navigating extensive rubrics. Once trained, human evaluators can complete evaluations relatively quickly using this approach.

\paragraph{Elo evaluation.}
While the automation rate measures absolute project completion against the human baseline, the Elo metric captures the relative performance between different AI agents, combining project completion with overall quality. This allows models to eventually exceed the human Elo score of $1,\!000$. The Elo evaluation involves a pairwise comparison between two AI Deliverables (AD-$1$ and AD-$2$). We use a modified version of the evaluation platform that displays both AI deliverables, along with the human deliverable as a reference for successful completion.

Evaluators assess the comparison along two dimensions using separate $3$-point Likert scales:
\begin{itemize}
    \item \textbf{Project completion}: Which deliverable is closer to satisfying the brief (i.e., closer to a state where it would be accepted by a reasonable client)? (AD-$1$ closer / Equally close / AD-$2$ closer)
    \item \textbf{Overall quality}: Which deliverable has higher overall quality for the project? (AD-$1$ higher / Same quality / AD-$2$ higher)
\end{itemize}
To compute the Elo score, we derive a unified preference from these two dimensions. We prioritize the project completion judgment when at least one of the AI agents has failed to complete the project. If both agents have successfully completed the project, we switch to using the overall quality judgment.

\paragraph{Evaluation standards and statistics.}
In all evaluations, we instruct evaluators to adopt the perspective of a reasonable client to minimize subjectivity. This grounds quality assessments in the likely reception of the work in a professional context, rather than the evaluators' personal preferences. We use majority voting across three independent evaluations to determine the final judgment. For Elo evaluations, if the three evaluations are split across the $3$-way Likert scale (e.g., one vote for AD-$1$, one for AD-$2$, and one for a tie), this is recorded as indifference.

The evaluation process demonstrates high reliability, with an inter-annotator agreement of $94.4\%$ for the automation rate metric. For Elo evaluations, ternary inter-annotator agreement is $56.9\%$, far above random chance of $33.0\%$. The probability of hard disagreements (one vote for AD-$1$ and one vote for AD-$2$) is $5.9\%$, indicating that evaluators are directionally nearly always in agreement.

Evaluation times are shown in \Cref{fig:evaluation-times}. Evaluators were requested to take a maximum of $20$ minutes for Automation Rate evaluations and $30$ minutes for Elo evaluations. These times were selected based on preliminary testing and provided ample time for completing most evaluations. Evaluations took $11.4$ minutes on average for Automation Rate and $17.4$ minutes for Elo. We hypothesize that the automation rate inter-annotator agreement rate will fall as AI deliverables become more complex, which could be countered with more experienced evaluators and longer evaluation time.
\begin{table}[t]
\captionsetup{skip=10pt}
\centering
\begin{tabular}{lc}\toprule
Model & Automation Rate \\\midrule
Manus & 2.5\% \\
Grok 4 & 2.1\% \\
Sonnet 4.5 & 2.1\% \\
GPT-5 & 1.7\% \\
ChatGPT agent & 1.3\% \\
Gemini 2.5 Pro & 0.8\% \\
\bottomrule
\end{tabular}
\caption{Current AI agents perform near the floor on RLI, solving less than $3\%$ of tasks in the benchmark.}\label{tab:main-table}
\vspace{-10pt}
\end{table}

\section{Experiments}

We evaluate the performance of several frontier AI agents on the Remote Labor Index (RLI) to assess the current state of AI automation capabilities on diverse economically valuable projects. We detail our experimental setup (Section \ref{sec:setup}), present quantitative results measuring both absolute and relative performance (Section \ref{sec:quantitative-results}), and provide a qualitative analysis of observed failure modes and agent behaviors (Section \ref{sec:qualitative-findings}).

\subsection{Experimental Setup}
\label{sec:setup}

\paragraph{Models and Environments.} We evaluate six state-of-the-art AI agents: ChatGPT agent \citep{chatgpt-agent}, GPT-5 \citep{gpt_5}, Claude Sonnet 4.5 \citep{sonnet_4_5}, Grok 4 \citep{grok4}, Gemini 2.5 Pro \citep{gemini-2.5-pro}, and  Manus \citep{manus}. For models that support computer-use, we used a computer-use scaffold developed by Scale AI. For models that do not support computer-use, we use the OpenHands scaffold, which we refer to as a command line interface (CLI) environment as opposed to a computer-use agent (CUA) environment. For GPT-5, we evaluated both the CUA and CLI scaffolds and report the CLI scaffold in the main tables, as this outperformed the CUA scaffold for this model. A full comparison of performance across environments is available in Appendix \ref{app:full-results}.

\paragraph{Scaffolding and prompting.} To ensure a fair assessment of peak capabilities, we tune prompts and provide standardized tooling scaffolds. This includes equipping agents with necessary execution tools and providing clear instructions on interfacing with the evaluation platform. For comprehensive details on the experimental setup, including the full prompts used, see Appendix \ref{app:evaluation_details}.

\begin{figure}[t]
\centering
\includegraphics[width=0.8\linewidth]{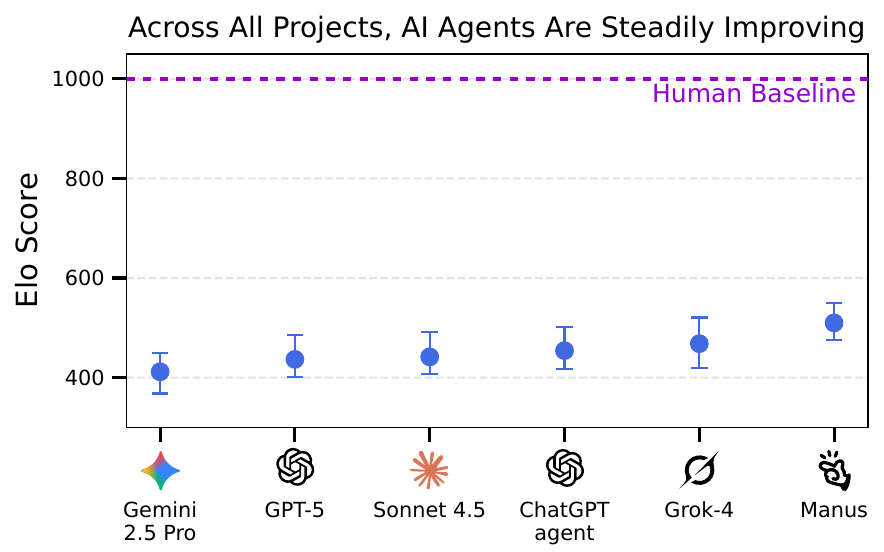}
\caption{Relative performance (Elo) scores show that AI agents are making steady progress on RLI and there are meaningful differences between models, despite all models falling short of the human baseline of $1,\!000$. Compared to the automation rate metric, Elo score provides a better measure of partial progress across all projects, including projects that are not solved yet.}
\label{fig:elo-rankings}
\end{figure}

\subsection{Quantitative Results}
\label{sec:quantitative-results}

We analyze the performance of AI agents on RLI using both absolute metrics (measuring success against the human baseline) and relative metrics (measuring progress between models). The main results are summarized in \Cref{tab:main-table}.

\paragraph{Absolute performance is near the floor.}
The central finding of our evaluation is that current AI agents demonstrate minimal capability to perform the economically valuable projects in RLI. We measure this capacity using the Automation Rate: the percentage of projects completed at a quality level equivalent to or exceeding the human gold standard. Across all models evaluated, absolute performance is near the floor, with the highest Automation Rate achieved being only $2.5\%$ (Manus).

Correspondingly, the metrics tracking the economic impact of automation (Dollars Earned and Autoflation) are also close to the floor. These results indicate that contemporary AI systems fail to complete the vast majority of projects at a level that would be accepted as commissioned work in a realistic freelancing environment. Despite rapid progress on other AI benchmarks, current systems remain far from capable of autonomously handling the diverse and complex demands of the remote labor market.

\paragraph{Elo score reveals steady improvement.}
While absolute performance remains low, it is crucial to detect more granular signs of progress. To measure the relative performance between different models, we use pairwise comparisons to compute an Elo score that represents how close models are to completing projects along with the overall quality of their deliverables. This enables tracking improvements between models, even when they fail to fully complete most projects.

We find that progress is measurable on RLI. The Elo rankings (Figure \ref{fig:elo-rankings}) indicate that models are steadily improving relative to each other, and the rankings generally reflect that newer frontier models achieve higher performance than older ones. This demonstrates that RLI is sensitive enough to detect ongoing progress in AI capabilities.

\subsection{Qualitative Findings}
\label{sec:qualitative-findings}

To understand the limitations of current systems and the reasons for the low automation rates, we conducted a qualitative analysis of agent failures by clustering the written justifications provided by evaluators. This analysis reveals a variety of failure modes, ranging from general quality issues to common systematic errors.

\paragraph{Common failure modes.}
Our qualitative analysis across roughly $400$ evaluations shows that rejections predominantly cluster around the following primary categories of failure:

\begin{enumerate}
    \item \textbf{Technical and File Integrity Issues}: Many failures were due to basic technical problems, such as producing corrupt or empty files, or delivering work in incorrect or unusable formats.
    \item \textbf{Incomplete or Malformed Deliverables}: Agents frequently submitted incomplete work, characterized by missing components, truncated videos, or absent source assets.
    \item \textbf{Quality Issues}: Even when agents produce a complete deliverable, the quality of the work is frequently poor and does not meet professional standards.
    \item \textbf{Inconsistencies}: Especially when using AI generation tools, the AI work often shows inconsistencies between deliverable files.
\end{enumerate}

\begin{wraptable}{r}{0.5\textwidth}
    \captionsetup{skip=10pt}
    \centering
    \begin{tabular}{lc}
        \toprule
         & Frequency (\%) \\
        \midrule
        Corrupted files & $17.6$ \\
        Incomplete & $35.7$ \\
        Poor quality & $45.6$ \\
        Inconsistencies & $14.8$ \\
        \bottomrule
    \end{tabular}
    \caption{Percentage of AI deliverables exhibiting issues, by category. Categories are not mutually exclusive; a deliverable may be counted in multiple categories.}
    \label{tab:qualitative_stats}
\end{wraptable}

For each AI deliverable we assigned one or more failure categories based on issues observed during the evaluations. Table \ref{tab:qualitative_stats} reports the proportion of deliverables affected by each category. Representative failure modes include: videos far shorter than requested (e.g., 8 seconds rather than 8 minutes), child-like drawings using basic geometric shapes, inconsistent visual appearance across renderings (e.g., a house's appearance changing across different 3D views), robotic or unnatural voice-overs, digital floor plans that do not match the supplied sketches, and web games that function but whose graphics fall short of professional standards.

\paragraph{Successful AI deliverables.} Across a small subset of projects, AI deliverables were judged comparable or better than human output. These were predominantly creative projects, especially audio and image related work, along with writing and data retrieval/web scraping. Specifically, across all models we tested, performance matched or exceeded human baselines on several audio editing, mixing and production tasks (e.g., creating bespoke sounds effects for a retro video game, separating vocals from accompaniment in a single track, merging voice-overs with intro and outro music) and on image-generation tasks (e.g., ad and logo creation). AI also performed well on report writing and on generating code for interactive data visualization. We provide examples of successful and unsuccessful AI deliverables (see Appendix \ref{app:ai-examples}).

\paragraph{Cognitive skills analysis.}
Hendrycks et al. \citep{hendrycks2025agidefinition} show that the skills and weaknesses of LLMs can be decomposed into several distinct categories, such as broad world knowledge, memory, and audiovisual abilities. We observe that many of the failures exhibited by AI agents stem from deficits in these skills. For example, many failures stem from AI agents being unable to verify the correctness of their work and fix mistakes, especially in projects requiring complex and interactive audiovisual verification, such as architecture, game development, and web development. Analogously, many of the successes of AI models lie in domains where current AI models' skills are more developed, such as projects where the complexity is primarily in text processing or image creation.

\begin{figure}[t]
\vspace{-10pt}
\centering
\includegraphics[width=1\linewidth]{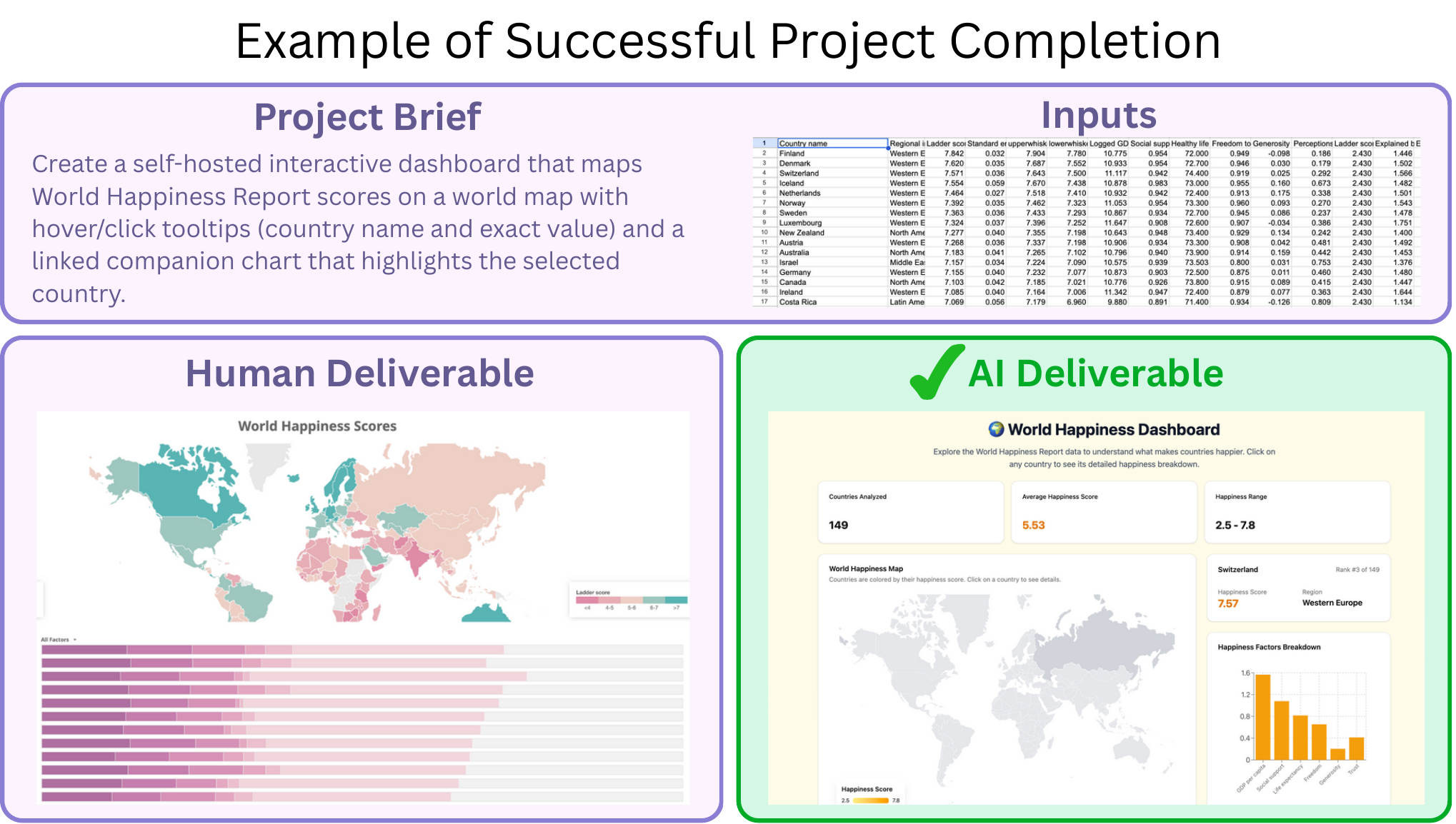}
\caption{Here we show a successful project completion from Sonnet 4.5. Simple web visualizations that only require writing code are well within the capabilities of current AI agents, but this work makes up a small slice of all remote labor. Additional examples of successes and failures are shown in \Cref{fig:ai-deliverable-examples-1,fig:ai-deliverable-examples-2}.}
\label{fig:successful-deliverable-main-paper}
\end{figure}

\section{Discussion}

\paragraph{Generalization to automating new jobs.}

Historically, automation technologies have been task-specific: the electronic calculator automated the job of human calculators, but when these workers re-trained and focused on skills that had not yet been automated, the calculator wasn't able to automate any of these new tasks. This is because humans have general cognitive skills that calculators do not.

AI differs qualitatively from other automation technologies; it is not designed merely to automate specific tasks, but is being explicitly developed to automate human intelligence itself. Indeed, current AIs are not task-specific, but rather have general cognitive skills and are already capturing a substantial fraction of human-level cognitive generality \citep{hendrycks2025agidefinition}. An AI that automates all current remote work without overfitting is likely to have many of the same general cognitive skills as humans, allowing it to automate new jobs as they arise \citep{korinek2024scenarios}. In this way, AIs may prove qualitatively different from prior automation technologies. While RLI does not fully represent every part of the remote labor economy, it is a substantial step towards measuring the ability of AI to automate the remote economy in general, rather than just current tasks.

\paragraph{Limitations.}

RLI excludes some types of work found commonly in the remote labor economy, including projects requiring interaction with the client (e.g. tutoring), jobs that require working on a team (e.g., project management), and other types of work that did not meet our requirements (see Appendix \ref{app:task_filtering} for the full list of requirements). While RLI is the broadest benchmark of its kind, it does not represent several types of remote work due to these constraints. Thus, an AI obtaining $100\%$ automation rate on RLI may still underperform humans on types of work that we do not evaluate.

The cost of the projects reported by human professionals reflects the cost at the time of project completion and is not adjusted for inflation. In most cases where we know the project completion date, the projects were completed in the past five years; consequently, the reported costs likely underestimate the current economic value of this work when accounting for inflation.

\section{Conclusion}

RLI establishes an economically grounded measure of AI automation capacity, with $240$ projects spanning $23$ domains of digital freelance work, each anchored in demonstrated market value. Frontier AI agents perform near the floor on RLI, achieving an automation rate of less than $3\%$, revealing a stark gap between progress on computer use evaluations and the ability to perform real and economically valuable work. RLI aims to establish the empirical foundation stakeholders need to monitor AI capabilities, forecast labor market impacts, and proactively navigate AI-driven automation.

\section*{Acknowledgments}
We would like to thank Anders Edson, Hale Guyer and Connor Smith for providing helpful feedback throughout the drafting process. We would also like to thank Michael Jae Byun and Brian Jang for helpful discussions.

\bibliography{main}
\bibliographystyle{plain}

\appendix

\newpage

\begin{table}[t]
\captionsetup{skip=10pt}
\centering

\begin{minipage}[t]{0.48\textwidth}
\centering
\begin{tabular}{lc}\toprule
Model & Automation Rate \\\midrule
Manus & 2.5\% \\
Grok 4 & 2.1\% \\
Sonnet 4.5 & 2.1\% \\
GPT-5 (CLI) & 1.7\% \\
ChatGPT agent & 1.3\% \\
GPT-5 (CUA) & 0.8\% \\
Gemini 2.5 Pro & 0.8\% \\
\bottomrule
\end{tabular}
\end{minipage}\hfill
\begin{minipage}[t]{0.48\textwidth}
\centering
\begin{tabular}{lc}\toprule
Model & Elo \\\midrule
Manus & 509.9 \\
Grok 4 & 468.2 \\
ChatGPT Agent & 454.3 \\
Sonnet 4.5 & 441.7 \\
GPT-5 (CLI) & 436.7 \\
GPT-5 (CUA) & 431.6 \\
Gemini 2.5 Pro & 411.8 \\
\bottomrule
\end{tabular}
\end{minipage}

\caption{Full automation rate and Elo results. In Appendix \ref{app:agent-environments}, we describe our comparison of two agent scaffolds for GPT-5, a command-line interface (CLI) scaffold and computer-use (CUA) scaffold. In the main paper, we show GPT-5 with the CLI scaffold.}
\label{tab:full-automation-elo}
\end{table}

\newcolumntype{/}{D{/}{/}{-1}} 

\begin{table}[t]
\captionsetup{skip=10pt}
\centering
\begin{tabular}{l/}\toprule
Model & \text{Dollars Earned} / \text{Max Possible} \\\midrule
Manus & \$1,\!720 / \$143,\!991 \\
Sonnet 4.5 & \$1,\!280 / \$143,\!991 \\
GPT-5 (CLI) & \$1,\!180 / \$143,\!991 \\
Grok 4 & \$858 / \$143,\!991 \\
GPT-5 (CUA) & \$858 / \$143,\!991 \\
ChatGPT agent & \$520 / \$143,\!991 \\
Gemini 2.5 Pro & \$210 / \$143,\!991 \\
\bottomrule
\end{tabular}
\caption{Current models earn a small fraction of the total cost of projects in the dataset.}
\label{tab:dollars-earned}
\end{table}

\section{Additional Results}

\subsection{Full Results}
\label{app:full-results}

In \Cref{tab:full-automation-elo}, we show the precise Elo score and automation rate for all models, including the CLI and CUA scaffolds for GPT-5.

In \Cref{tab:dollars-earned}, we show the dollars earned for all evaluated models. Current AI agents earn a small fraction of the total cost of projects in the dataset.

\subsection{Autoflation}
\label{app:autoflation}

\begin{figure}[t]
  \centering
  \includegraphics[width=\linewidth]{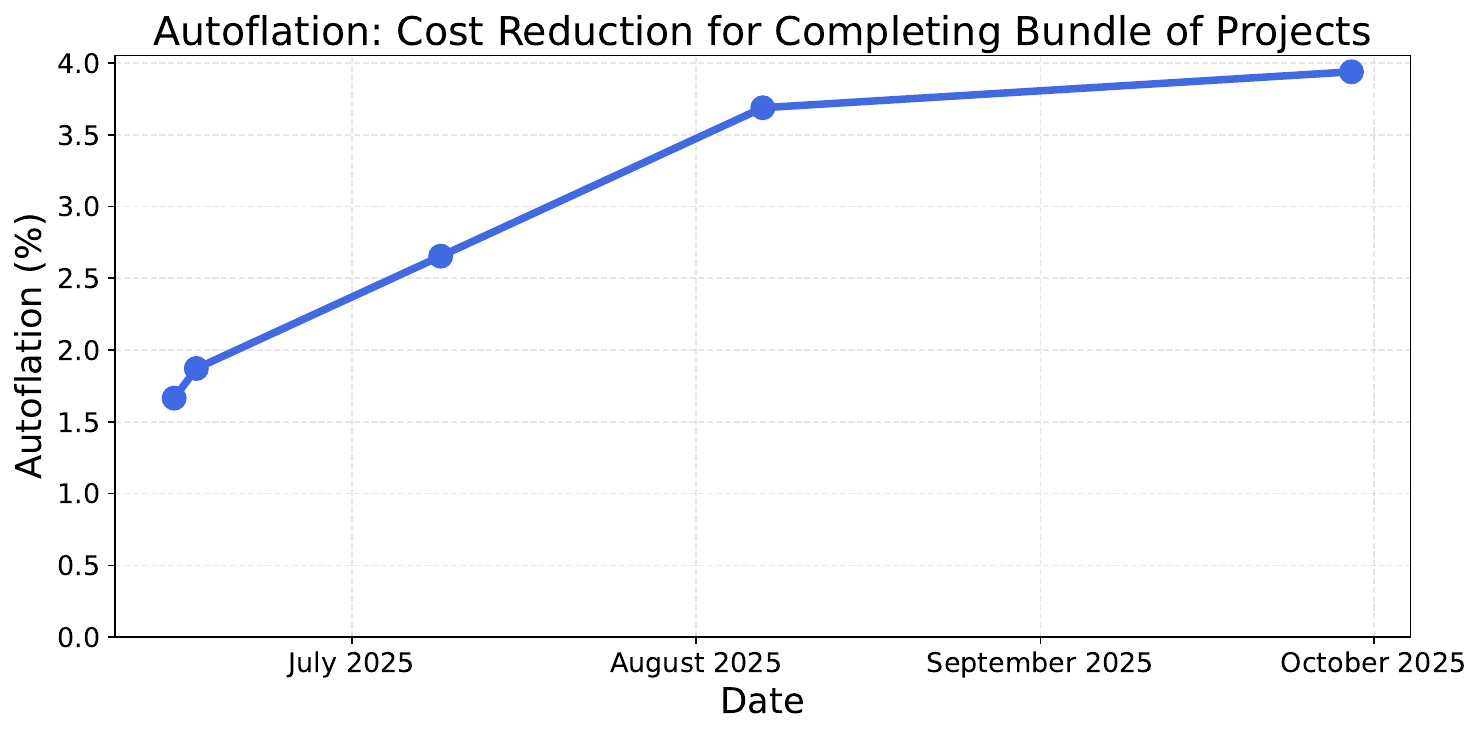}
  \caption{Autoflation on RLI: the percentage decrease in the cost of completing the fixed RLI project bundle, using AI agents to complete projects if they successfully complete them at lower cost than humans. As AI systems achieve the same deliverables at lower effective cost, the price of this work declines.}
  \label{fig:autoflation-frontier}
\end{figure}

In \Cref{fig:autoflation-frontier}, we show the reduction in the cost of completing the projects in RLI. Analogous to indices that track the price of bundles of goods, this lets us track deflation in the effective price of the fixed bundle of projects represented by RLI. We refer to this quantity as ``autoflation'' and plot how it changes over time as new models are released.

For each project, we measure the cost difference relative to the human-produced deliverable when using the lowest-cost method of achieving an acceptable deliverable. If no AI method completes the project at a lower effective cost than the human baseline, the reduction is zero for that project. Because the metric is sensitive to false positives in annotation, we audit all AI deliverables marked as successful to minimize the false-positive rate.

\subsection{Effect of Agent Scaffolds}\label{app:agent-environments}

Our results suggest that current models are not yet able to take full advantage of computer-use environments. For instance, GPT-5 demonstrated superior performance when using a CLI-based agent compared to the Computer-Use Agent (CUA) setup. This holds for both the Elo scores (CLI: $436.7$; CUA: $431.6$) and the automation rates (CLI: $1.7\%$; CUA: $0.8\%$). We expect more vertical integration of model scaffolds will yield stronger performance.

\section{Evaluation Details}\label{app:evaluation_details}

\subsection{Model Details}

The vast majority of Manus deliverables were generated over the course of June, 2025. Some deliverables were generated in September, 2025.

Our Gemini evaluations are with Gemini 2.5 Pro, not Gemini 2.5 Computer Use. We found that the latter struggled with our computer-use environment, since it was tuned to work with browser-only environments.

\subsection{Elo Computation}

\paragraph{Collecting preference data.}
Projects and model pairs are randomly sampled for comparison, using random ordering of model pairs to remove order effects. We use stratified sampling across models to ensure each model pair is compared on at least $10$ projects (median $25$). These are combined with the automation rate evaluations (model vs human) to obtain the final preference data.

For each project that a model pair is compared on, we perform majority voting, using two independent evaluations with a third to break ties if needed. In cases where the three evaluations are ``prefer AD-1'', ``indifferent'', and ``prefer AD-2'', we code the preference as indifference on this project. In cases where the majority vote is for indifference, we code the preference as $50/50$. We numerically average these preferences across all compared projects to obtain a probabilistic preference for the model pair. These probabilistic preferences make up the preference graph.

\paragraph{Fitting Bradley-Terry utilities.}
Following the Chatbot Arena methodology \citep{chiang2024chatbot}, we use global Bradley-Terry fitting on sampled preference edges to compute utility scores, which we refer to as Elo scores for ease of understanding. We use $100$ bootstrap samples to compute $95\%$ confidence intervals in \Cref{fig:elo-rankings}. Bootstrap samples are taken over projects, followed by re-averaging preferences on the sampled projects to obtain probabilistic preferences.

\paragraph{Normalizing scores.}
After computing Bradley-Terry utilities, we scale and shift the utilities so that the human baseline obtains a score of $1,\!000$ and a difference in score of $400$ corresponds to $10\!:\!1$ odds of winning.

\subsection{Evaluation and Generation Budgets}

\begin{figure}[t]
\centering
\includegraphics[width=1.0\linewidth]{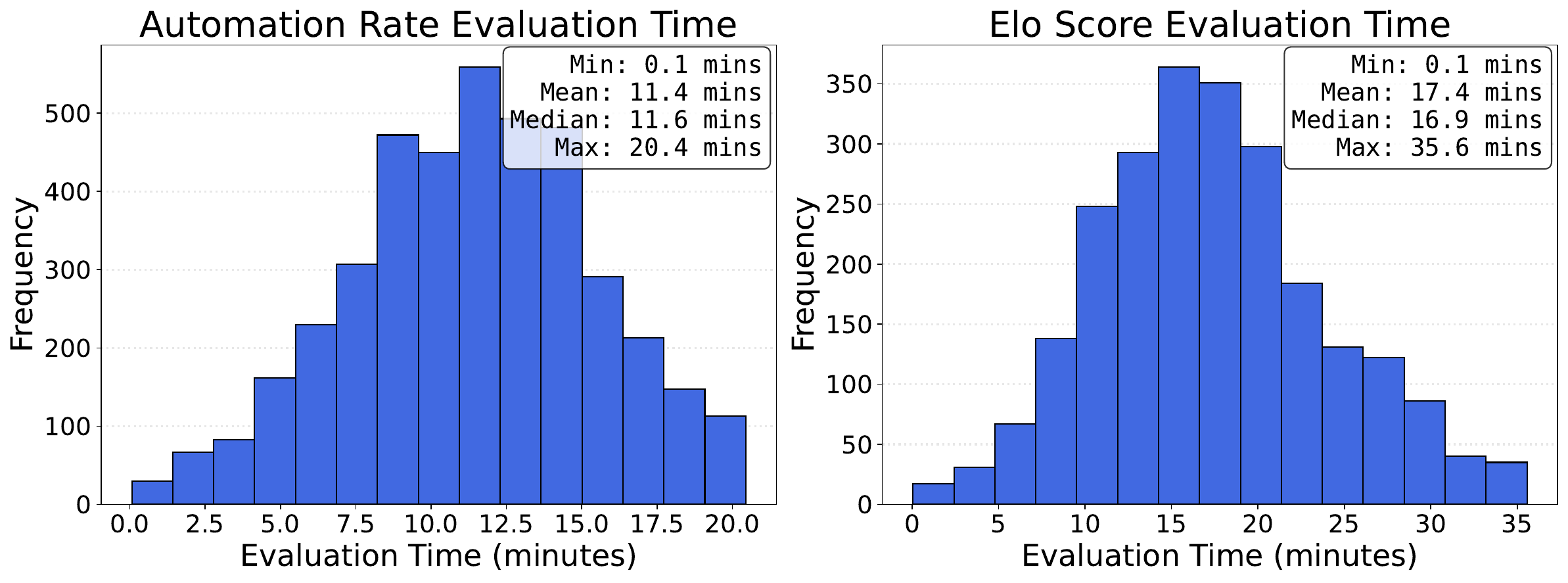}
\caption{Each evaluator was given a soft maximum of $20$ minutes for model vs human evaluations and $30$ minutes for model vs model evaluations (the latter requires inspecting more files and takes more time). In preliminary testing, we found this duration was adequate for nearly all projects. Total evaluation time per project is higher, as $2$ to $3$ evaluations were performed to obtain a majority vote.}
\label{fig:evaluation-times}
\end{figure}

\paragraph{Evaluation time budget.}
Evaluators were asked to spend no more than $20$ minutes per project for automation rate evaluations and no more than $30$ minutes per project for Elo evaluations. These times were selected based on preliminary testing and provided ample time for completing most evaluations. Elo evaluations involve inspecting two AI deliverables, and hence require more time. As shown in \Cref{fig:evaluation-times}, most evaluations finished in less than this amount of time, with a small number exceeding it.

For current AI agents, evaluations are possible to complete relatively quickly, because AI deliverables often have glaring errors that are easy to spot. As AI deliverables become more complex and come closer to solving the projects in RLI, we expect that the time needed for evaluating each project will increase.

\subsection{Evaluation Instructions}\label{app:human_evaluator_instruction}

\paragraph{Evaluator training materials.}

Before beginning evaluations, annotators were required to review detailed instructional videos and documents covering the evaluation workflow and common pitfalls. The training emphasized three core principles for evaluation:

\begin{itemize} \item \textbf{Reasonable Client Perspective:} We instructed annotators to judge each deliverable holistically from the perspective of a reasonable client commissioning the project. This approach grounds quality assessments in the likely reception of the work in a professional context, minimizing the evaluators' personal subjectivity. \item \textbf{Zone of Acceptable Error:} The human reference deliverable establishes the baseline level and quality of work accepted by the original client. Annotators were instructed to view the reference within a zone of acceptable error - if the human deliverable contained minor flaws or was missing non-critical components, the AI deliverable was held to the same standard and not penalized for similar omissions. \item \textbf{Common AI Failure Modes:} The training materials highlighted specific, common issues prevalent in AI-generated work. Examples included the use of rasterized image generation for projects explicitly requiring vector graphics, the inclusion of unreadable or nonsensical text in images, and a lack of spatial or visual consistency across different files within the same deliverable. \end{itemize}

The training detailed the standardized evaluation workflow: Annotators must first gain an understanding of the project by reading the brief and reviewing the reference deliverable. With this baseline established, they evaluate the AI deliverable(s) based on the requirements of the brief and the human reference. Annotators were allotted time limits for evaluation ($20$ minutes for Human vs. Model; $30$ minutes for Model vs. Model). However, they were instructed to stop early and fail the project if they identified a critical flaw that rendered the deliverable unusable. We iterated on these evaluation instructions and audited annotator quality until achieving an inter-annotator agreement of $\geq$85\% on a random subset of the projects. Our final version of the instructions achieved $94.4\%$ inter-annotator agreement.

\paragraph{Automation Rate Evaluation Instructions.}

In the Automation Rate evaluation, evaluators assess the AI deliverable (``alternative deliverable'' or ``AD'') using the human deliverable as a reference for what successful project completion looks like (``reference deliverable'' or ``RD''). After reviewing the project materials according to the trained workflow, evaluators must provide a classification based on the following 3-point scale, accompanied by a written justification:

\begin{enumerate}
    \item The alternative deliverable does not satisfy the brief as well as the reference deliverable or is of significantly lower quality, such that it would not be accepted by a reasonable client as the commissioned work.
    \item The alternative deliverable satisfies the brief as well as the reference deliverable and would be accepted by a reasonable client as the commissioned work.
    \item Same as 2, and the alternative deliverable exceeds the reference deliverable in overall quality.    
\end{enumerate}

The automation rate is calculated based on the percentage of projects receiving a rating of 2 or 3. The distinction between equal (2) and superior (3) quality is maintained to facilitate Elo computations and may help provide greater clarity into the abilities of models near human parity in the future.

\paragraph{Elo Score Evaluation Instructions.}
The Elo evaluation involves a pairwise comparison between two AI deliverables (AD-1 and AD-2). The evaluation platform displays both AI deliverables, along with the human deliverable as a reference for what successful project completion looks like. Annotators assess the comparison along two dimensions using separate 3-point scales:

Project completion:
\begin{enumerate}
  \item AD-1 is closer to satisfying the brief than AD-2, meaning AD-1 is closer to a state where it would be accepted by a reasonable client as the commissioned work.
  \item AD-1 is equally close to satisfying the brief as AD-2, meaning both are equally close to a state where they would be accepted by a reasonable client as the commissioned work.
  \item AD-2 is closer to satisfying the brief than AD-1, meaning AD-2 is closer to a state where it would be accepted by a reasonable client as the commissioned work.
\end{enumerate}

Overall quality:
\begin{enumerate}
  \item AD-1 has higher overall quality for the project than AD-2.
  \item AD-1 has the same overall quality for the project as AD-2.
  \item AD-2 has higher overall quality for the project than AD-1.
\end{enumerate}

\subsection{Evaluation verification}

To reduce the rate of false positives, we manually audited all annotation cases where the AI deliverable was labeled as good or better than the human deliverable. We were able to audit all of those cases since there were only a small number of these annotations. 

To get a false negative rate, two co-authors randomly sampled a Human vs Model pair from $50$ random projects and did manual evaluation on those projects. We found no false negatives (cases where a annotators incorrectly labeled the human deliverable to be preferable). This gives us $\leq 5.8\%$ false negative rate with $95\%$ confidence.

\subsection{Agent Setup}\label{app:model_prompts}

\paragraph{Scaffolds.}
We use three types of agent scaffold:
\begin{itemize}
  \item Integrated agents (ChatGPT agent, Manus)
  \item Computer-use environment developed by Scale AI
  \item OpenHands (CLI-based environment)
\end{itemize}

For models that support computer-use (GPT-5, Sonnet 4.5), we default to our computer-use scaffold. For models not supporting computer-use (Grok 4, Gemini 2.5 Pro), we use OpenHands.

In Appendix \ref{app:agent-environments}, we compare GPT-5 on both OpenHands and computer-use scaffolds. We refer to these as GPT-5 (CLI) and GPT-5 (CUA), respectively. In the main results, we report GPT-5 (CLI), as this outperformed GPT-5 (CUA).

\begin{figure}
    \centering
    \includegraphics[width=0.5\linewidth]{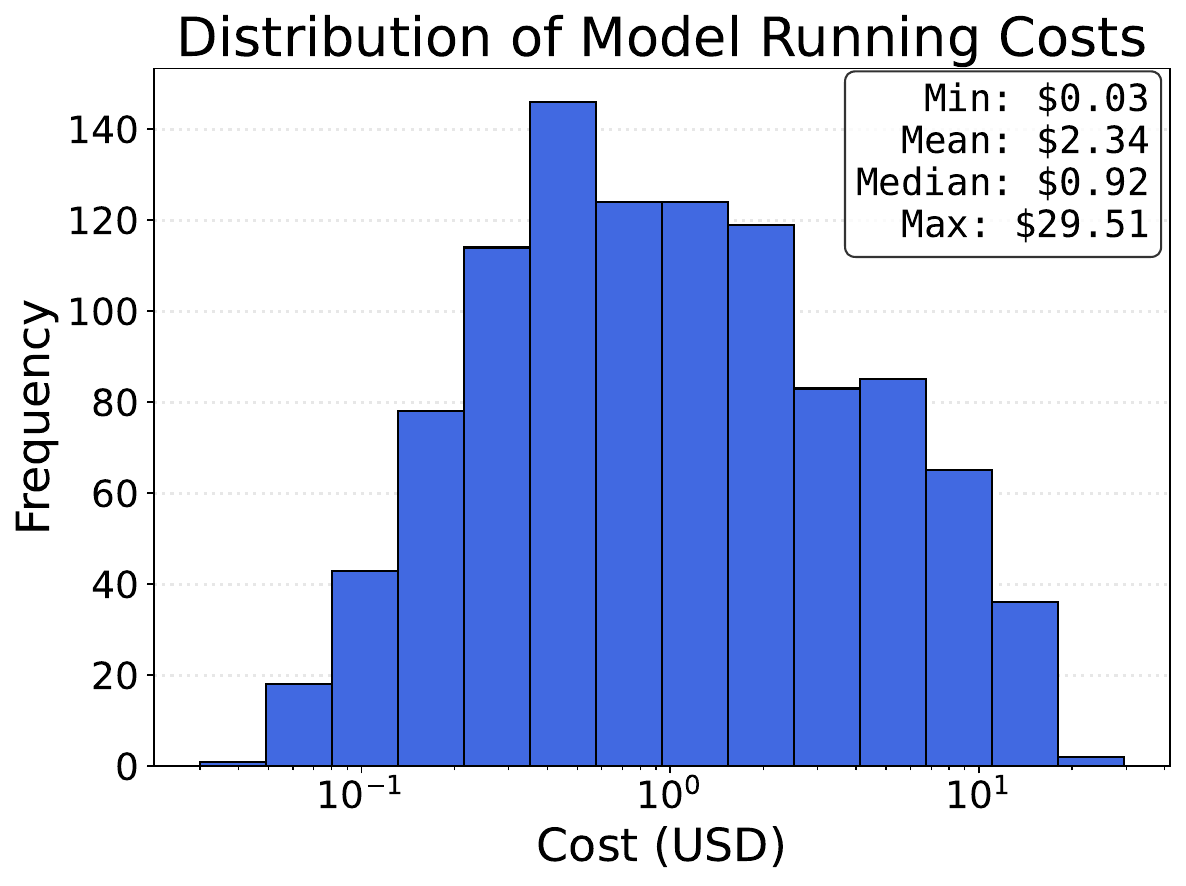}
    \caption{The average cost of generating AI deliverables was $\$2.34$. In all cases, models stopped generating before exceeding $\$30$ of API costs.}
    \label{fig:generation_cost_distribution}
\end{figure}

\paragraph{Available software and tools.}

To support the diverse multimedia requirements of RLI projects, we augmented the OpenHands environment and gave the models access to the following tools: gpt-image-1 for image generation and editing, openai/tts-1 for speech generation, and veo-3.0-generate-preview for video generation. The agent scaffolding included standardized instructions for managing input and deliverable directories and explicitly directed agents to prioritize using these specialized tools over general LLM queries when the required functionality was available.

Our computer use agent was deployed in remote Ubuntu VM instances via the Scrapybara API and given access to three tool categories through the Model Control Protocol (MCP): computer use tools for mouse, keyboard, and screenshot operations, file editor tools for viewing, creating, and modifying files with string replacement and line insertion capabilities, and bash tools for shell command execution with persistent session state. The agent scaffolding employed a modular action system that converted high-level tool calls into standardized operations executed by interface handlers. To generate artifacts, agents received the project brief and input files, were directed to save deliverables to designated directories, and operated within a session timeout (default: 1 hour), with execution traces capturing all actions and outputs for evaluation.

\paragraph{Prompts.}

For all artifact generation setups with the exception of the computer use agent, we used the following base prompt:

\noindent
\begin{promptblock}
Read the brief attached and create only the deliverables described. Do not ask any questions. Complete the task and send a download link to the deliverables. You are done once all the deliverables are ready and the download link is sent.There may be auxiliary information necessary to complete the task that is provided in a zipped ``inputs'' folder. If this is provided, unzip the folder first and then proceed with completing the task. Make a zip file with all the deliverables.
\end{promptblock}

\vspace{10pt}
Additionally, we provided the models with specifications regarding the file types supported by the evaluation platform (see Appendix~\ref{app:evaluation_platform_details}), which accommodates a comprehensive range of file formats with fallback rendering capabilities for unsupported types. For Manus and ChatGPT Agent, no further scaffolding instructions were added.

For OpenHands, we extended the base prompt with instructions directing agents to use input/output directories and provided access to specialized multimedia tools:

\begin{promptblock}
NOTE: You can explore './inputs' directory for extra information and reference material to execute the task. The folder might be empty, meaning that no further information is provided.

IMPORTANT: Always save your final deliverables to the './output' directory. This directory has been created for you. Only put the requested deliverable output files in the './output' folder and no other extraneous files (eg. README's, etc.). Each deliverable file must also have an appropriate extension (eg. .jpg, .png, .pdf, .csv, etc.). You can save your intermediate scripts or files to the './auxiliary' directory but this is not required.

SPECIALIZED TOOLS: The './tools' directory contains specialized tools you can use to complete your tasks. These include:
- 'gpt-image-1': Image generation and editing
- 'openai/tts-1': Speech generation  
- 'veo-3.0-generate-preview': Video generation

You should absolutely use these tools if their functionality is needed to complete the task (instead of defaulting to general LLM query). Before using any tool, make sure to read its documentation and install any required dependencies. After execution, wait at least 300 seconds before killing the operation.
\end{promptblock}

\vspace{10pt}
For the computer use agent, we used the following prompt.

\noindent
\begin{promptblock}
Read the brief below and create only the deliverable described. Do not ask any questions. Try to work in /opt/workspace/ directory first, but if that's not accessible, work in the current directory. If you cannot find the inputs folder or get permission errors, call the navigate_to_workspace function first, then ensure_workspace_directories if needed. If you get ``Permission denied'' errors when saving files, call the fix_workspace_permissions function to resolve them. Complete the task, and make sure to submit all of the deliverables. You are done once all the deliverables are ready, and you have saved all deliverables to the Deliverables folder (either /opt/workspace/Deliverables/ or ./Deliverables/ depending on what's accessible). You are allowed to use temporary or auxiliary files, please save them in the auxiliary folder. Avoid long outputs when using bash, you can control the amount of output by using 'head' or 'tail' when using bash.
\end{promptblock}

\vspace{10pt}
For Claude Sonnet 4.5, we further extended the computer use agent prompt above with quality verification instructions to leverage the model's visual reasoning capabilities. Based on best-use recommendations suggested by early users of Claude Sonnet 4.5, we also implemented context management exceeding 1M tokens and included explicit instructions to verify any output code or files and to avoid excessively writing thinking traces to files.

\noindent
\begin{promptblock}
No need to write too many text file notes to the filesystem, try to keep your thoughts / reasoning / insights in your context window. Also verify that any outputs you generate (intermediate or final) are of good quality by taking screenshots of files for visual inspection and checking any code for potential errors.
\end{promptblock}

\subsection{Evaluation Platform Details}\label{app:evaluation_platform_details}

The evaluation platform is a web-based multimedia viewer and file explorer. It provides native support for viewing the following file types:

\begin{itemize}
    \item \textbf{Documents}:
    \begin{itemize}
        \item \textbf{Text}: \texttt{.txt}, \texttt{.json}, \texttt{.yml}, \texttt{.py}, \texttt{.js}, \texttt{.ts}, \texttt{.css}, \texttt{.java}, \texttt{.go}, \texttt{.php}, \texttt{.rb}, \texttt{.swift}, \texttt{.sql}, \texttt{.sh}, and other common source code files. Any non-binary file not otherwise supported is displayed as text.
        \item \textbf{Formatted}: \texttt{.md}, \texttt{.html}, \texttt{.pdf}, \texttt{.tex} (LaTeX), and \texttt{.ipynb} (Jupyter Notebooks).
        \item \textbf{Spreadsheets}: \texttt{.csv}, \texttt{.xls}, \texttt{.xlsx}.
        \item \textbf{Microsoft Office}: \texttt{.ppt}, \texttt{.pptx}, \texttt{.doc}, \texttt{.docx}.
    \end{itemize}
    \item \textbf{Media}:
    \begin{itemize}
        \item \textbf{Images}: \texttt{.jpg}, \texttt{.jpeg}, \texttt{.png}, \texttt{.gif}, \texttt{.bmp}, \texttt{.webp}, \texttt{.svg}, \texttt{.ico}, \texttt{.avif}, \texttt{.tif}, \texttt{.tiff}.
        \item \textbf{Video}: \texttt{.mp4}, \texttt{.m4v}, \texttt{.mkv}, \texttt{.webm}, \texttt{.mov}, \texttt{.avi}, \texttt{.wmv}.
        \item \textbf{Audio}: \texttt{.mp3}, \texttt{.wav}, \texttt{.ogg}, \texttt{.aac}, \texttt{.m4a}, \texttt{.midi}, \texttt{.mid}.
    \end{itemize}
    \item \textbf{Design \& 3D}:
    \begin{itemize}
        \item \textbf{Design}: \texttt{.psd} (with limited support for complex layer effects).
        \item \textbf{3D Models}: \texttt{.obj}, \texttt{.mtl}, \texttt{.stl}, \texttt{.gltf}, \texttt{.glb}.
        \item \textbf{Autodesk/CAD}: \texttt{.dwg}, \texttt{.dxf}, \texttt{.skp}, \texttt{.stp}, \texttt{.step}, \texttt{.ipt}, \texttt{.3dm}, \texttt{.3ds}, \texttt{.fbx}, \texttt{.rvt}, \texttt{.ifc}, and other formats supported by the Autodesk Viewer.
    \end{itemize}
    \item \textbf{Data \& Interactive}:
    \begin{itemize}
        \item \textbf{Databases}: \texttt{.sqlite}, \texttt{.db}.
        \item \textbf{Websites/WebGL}: Interactive builds with \texttt{.html} entry points and associated \texttt{.js} and \texttt{.css} assets.
        \item \textbf{Anki}: \texttt{.apkg} (limited to front and back card formats).
    \end{itemize}
\end{itemize}

The evaluation platform is fully open-source.

\begin{figure}
    \centering
    \includegraphics[width=1\linewidth]{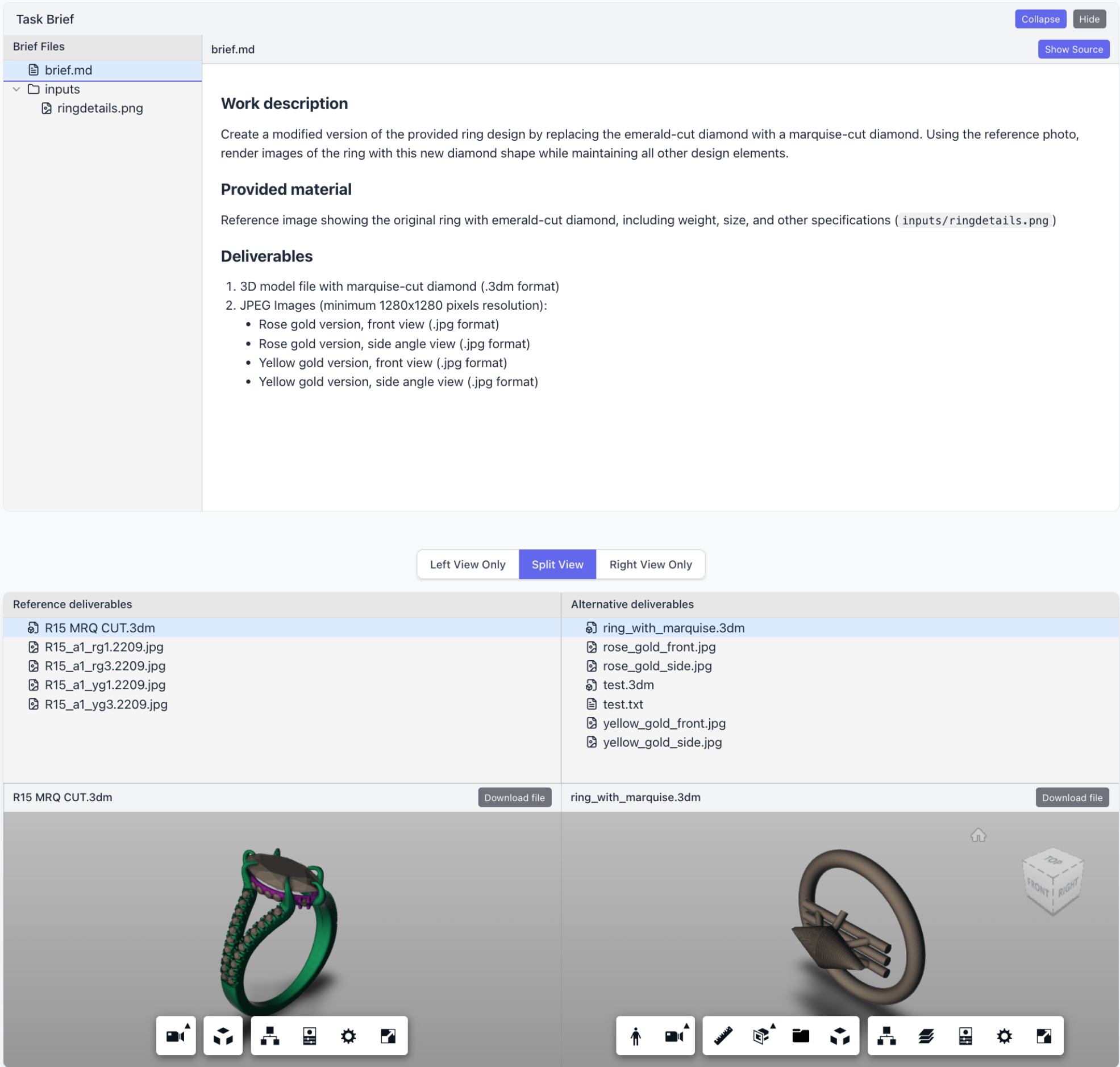}
    \label{fig:evaluation_platform}
    \caption{Evaluation platform view with the ring 3D model project example.}

\end{figure}

\paragraph{Project-specific notes for evaluation.}

For some projects, we display short notes in a popup in the evaluation platform. These evaluator notes contain project-specific details of how the evaluation should be performed. For example, in some projects the human deliverable contains additional features that we exclude from the project brief. In these cases, we instruct the evaluator to ignore those parts of the human deliverable and emphasize that the AI deliverable should not include those features. Less than $20$ projects have evaluator notes.

\begin{figure}[t]
  \centering
  \includegraphics[width=0.90\linewidth]{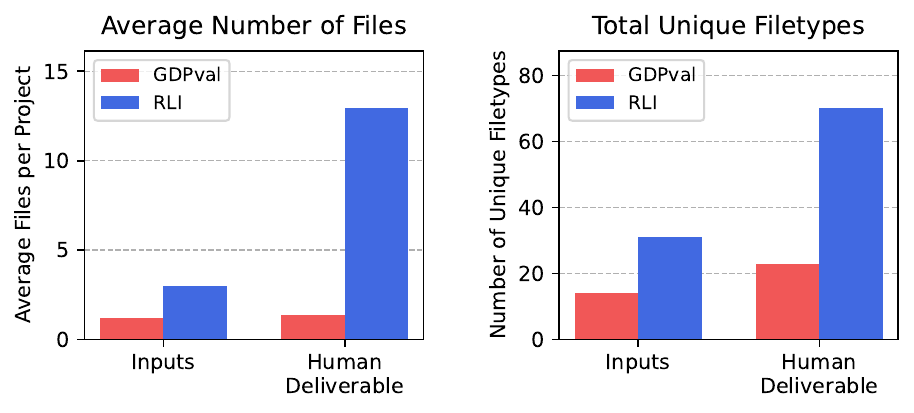}
  \caption{RLI projects involve significantly more diverse file types than previous comparable benchmarks. Left: Average number of files per project for inputs and human deliverables across benchmarks. Right: Total unique file types found in inputs and human deliverables across benchmarks.}
  \label{fig:filetype_comparison}
\end{figure}

\section{Dataset Details} \label{app:dataset_details}

\subsection{Categorization}
\label{app:upwork_domains}

\paragraph{Upwork taxonomy.}
We categorize all projects using the Upwork job taxonomy. We used the version current at the time of this paper's release, which contains $12$ major categories and $64$ subcategories of work. This taxonomy is detailed below.

\begin{itemize}
  \item \textbf{Accounting and Consulting}: Accounting \& Bookkeeping, Financial Planning, Management Consulting \& Analysis, Personal \& Professional Coaching, Recruiting \& Human Resources, Other - Accounting \& Consulting
  \item \textbf{Admin Support}: Data Entry \& Transcription Services, Market Research \& Product Reviews, Project Management, Virtual Assistance
  \item \textbf{Customer Service}: Community Management \& Tagging, Customer Service \& Tech Support
  \item \textbf{Data Science and Analytics}: AI \& Machine Learning, Data Analysis \& Testing, Data Extraction/ETL, Data Mining \& Management
  \item \textbf{Design and Creative}: Art \& Illustration, Audio \& Music Production, Branding \& Logo Design, Graphic, Editorial \& Presentation Design, NFT, AR/VR \& Game Art, Performing Arts, Photography, Product Design, Video \& Animation
  \item \textbf{Engineering and Architecture}: 3D Modeling \& CAD, Building \& Landscape Architecture, Chemical Engineering, Civil \& Structural Engineering, Contract Manufacturing, Electrical \& Electronic Engineering, Energy \& Mechanical Engineering, Interior \& Trade Show Design, Physical Sciences
  \item \textbf{IT and Networking}: Database Management \& Administration, DevOps \& Solution Architecture, ERP/CRM Software, Information Security \& Compliance, Network \& System Administration
  \item \textbf{Legal}: Corporate \& Contract Law, Finance \& Tax Law, International \& Immigration Law, Public Law
  \item \textbf{Sales and Marketing}: Digital Marketing, Lead Generation \& Telemarketing, Marketing, PR \& Brand Strategy
  \item \textbf{Translation}: Language Tutoring \& Interpretation, Translation \& Localization Services
  \item \textbf{Web, Mobile, and Software Development}: AI Apps \& Integration, Blockchain, NFT \& Cryptocurrency, Desktop Application Development, Ecommerce Development, Game Design \& Development, Mobile Development, Product Management \& Scrum, QA Testing, Scripts \& Utilities, Web \& Mobile Design, Web Development, Other - Software Development
  \item \textbf{Writing}: Content Writing, Editing \& Proofreading Services, Professional \& Business Writing, Sales \& Marketing Copywriting
\end{itemize}

Our final dataset includes projects from $9$ major categories and $23$ subcategories. In \Cref{fig:pie-chart}, we show the distribution across subcategories. For brevity, we use the following short-form names in the figure: ``Video'' for ``Video \& Animation'', ``CAD'' for ``3D Modeling \& CAD'', ``Graphic Design'' for ``Graphic, Editorial \& Presentation Design'', ``Game Dev'' for ``Game Design \& Development'', ``Audio'' for ``Audio \& Music Production'', and ``Architecture'' for ``Building \& Landscape Architecture''. To better reflect the diversity of projects, we separate out music composition projects into their own subcategory for the figure, as music composition differs considerably from other projects in Audio \& Music Production. Music composition projects make up roughly $6\%$ of the benchmark. Further subdivisions of this nature are possible, as most subcategories in the Upwork taxonomy consist of multiple distinct types of work, but for consistency we use the unmodified Upwork taxonomy for all other discussion in the paper.

Most of our analysis focuses on the subcategories in the Upwork taxonomy, so for brevity, we refer to these as ``categories'' in other parts of the paper.

\paragraph{O*NET taxonomy.} \label{app:onet_limitations}

The O*NET database \citep{onet-database} provides a widely used taxonomy of occupational requirements and work activities within the US labor market. While valuable for capturing activities performed in long-term occupations, it is not tailored to end-to-end freelance labor markets like Upwork, making it unsuitable for classifying RLI projects and estimating coverage. This limitation stems from O*NET's structure at both the activity and occupational levels.

To categorize a broad range of work, O*NET relies on an abstract hierarchy of Work Activities. Even the most granular taxonomy in O*NET, Detailed Work Activities (DWAs), does not provide meaningful granularity for measuring task breadth. The DWA taxonomy includes many ubiquitous and generic items such as ``Retrieve information from electronic sources,'' and ``Read materials to determine needed actions,'' \citep{onet_dwas}, and coverage of these DWAs does not indicate meaningful coverage of remote work task types. At the occupational level, O*NET classifications are designed to describe the broad, ongoing responsibilities of long-term workers. This structure does not align with the delivery of specific, self-contained freelance projects. For this reason, we use the Upwork taxonomy of remote freelance labor for coverage analysis, since this taxonomy is designed for categorizing freelance work.

\subsection{Filtering \& Cleaning Criteria}\label{app:task_filtering}

\paragraph{Project sourcing criteria.}
To enable building a high-quality standardized benchmark, we hired freelancers from categories on Upwork that met the following criteria:

\begin{enumerate}
    \item \textbf{Remote work:} It must be possible to complete projects without any physical labor (e.g., no local photography).
    \item \textbf{No open-ended jobs:} Most jobs in the category must be end-to-end projects that can be performed, not open-ended long-term contractor roles.
    \item \textbf{Can be completed independently:} The work can be completed independently by one freelancer and does not inherently require working on a team.
    \item \textbf{Does not require interaction with client:} The work does not inherently require interacting with clients (e.g., no tutoring).
    \item \textbf{Does not require interaction with client services:} The work does not require testing or interacting with live services set up by the client (e.g., no QA testing of client websites).
    \item \textbf{No scraping without permission:} The work does not involve scraping information from low-traffic websites or websites where bots are expressly forbidden.
    \item \textbf{Can be evaluated on the spot:} Some categories of work inherently require time to evaluate work outputs (e.g., SEO). These categories were excluded, ensuring that all projects can be evaluated on the spot. Note: This restriction does not apply to projects where evaluations take a long time but can still be performed on the spot.
    \item \textbf{Excluding certain categories:} Many projects in the Content Writing category can already be solved by AIs and would not provide much information to include. Thus, this category and related categories were excluded. (Note: These are category-level exclusions; individual projects from other categories were not excluded based on whether current models solved them.) Most legal categories were excluded due to PII concerns.
    \item \textbf{Renderability:} Deliverables must be possible to view in a web-based evaluation platform (e.g., no desktop application development).
\end{enumerate}

Based on these criteria, we entirely excluded projects from the following categories on Upwork during our initial project collection:

\textit{Personal \& Professional Coaching; Recruiting \& Human Resources; Project Management; Community Management \& Tagging; Customer Service \& Tech Support; Performing Arts; Photography; International \& Immigration Law; Public Law; Digital Marketing; Marketing, PR \& Brand Strategy; Desktop Application Development; Mobile Development; Product Management \& Scrum; QA Testing; Content Writing; Professional \& Business Writing; and Sales \& Marketing Copywriting.} 

This left us with $45$ total Upwork categories to source projects from. These sourcing criteria were also applied during long tail project collection.

\paragraph{Data cleaning and filtering.}
After receiving raw data, we conducted an extensive process of cleaning and filtering to ensure that all projects in the dataset met the following criteria:

\begin{enumerate}
    \item \textbf{Completeness:} The brief and input files are complete and sufficient, with no additional external information needed to complete the project.
    \item \textbf{Anonymization:} The input files and deliverables do not include sensitive personal information pertaining to the client. Client faces were blurred out, and company names and logos were replaced with fake alternatives that preserve the realism of projects.
    \item \textbf{Human deliverable completes the project:} The gold-standard human deliverable successfully completes the project, such that a reasonable client would accept it as the commissioned work. Note: The majority of projects in RLI were paid for by clients, so this is often guaranteed by default.
    \item \textbf{File quality:} Input files are high-quality. E.g., if the raw data for projects sourced from freelancers includes low-resolution images or screenshots, we request higher-quality replacements from freelancers.
    \item \textbf{Faithful to the raw data:} For projects sourced from freelancers, we ensure that the cleaned projects are as faithful as possible to the raw data sent by the freelancers, using similar or identical phrasing to original client requests where possible.
    \item \textbf{Standardized structure:} All projects are standardized to have briefs with three top-level sections: ``Work description'' describing the work to be done, ``Provided material'' describing the auxiliary project inputs, and ``Deliverables'' describing the expected deliverables.
    \item \textbf{Renderability:} We ensure that all inputs and human deliverables are viewable in the evaluation platform. We convert unsupported formats to supported ones (e.g., AI to layered PDF) and exclude projects that cannot be supported. This often required improving the capabilities of the evaluation platform to accommodate projects with new file types.
\end{enumerate}

After the cleaning and filtering process, the dataset contains $240$ projects from the following $23$ Upwork subcategories:

\textit{Video \& Animation, 3D Modeling \& CAD, Graphic \& Editorial Design, Audio \& Music Production, Building \& Landscape Architecture, Product Design, NFT, AR/VR \& Game Art, Art \& Illustration, Interior \& Trade Show Design, Web Development, Branding \& Logo Design, Game Design \& Development, Management Consulting \& Analysis, Data Entry \& Transcription Services, Data Analysis \& Testing, Language Tutoring \& Interpretation, Data Extraction/ETL, Presentation Design, Web \& Mobile Design, Corporate \& Contract Law, Translation \& Localization Services, Market Research \& Product Reviews.}

\subsection{Analysis Details}

\paragraph{Completion time comparison.}

In \Cref{fig:upwork_comparison}, we extracted completion time data from the papers for GDPval \citep{patwardhan2025gdpval} and HCAST \citep{rein2025hcast}. To determine the average completion time and cost for Upwork projects, we analyzed $275$ completed jobs from a random sample of $60$ freelancers, using the hours worked and dollars earned for each job.

\paragraph{Project type comparison.}

In \Cref{fig:upwork_comparison}, we computed the distribution over project types for RLI and GDPval by using a judge LLM to classify the project briefs using the following instructions.

\begin{promptblock}
Classify this task into one of three categories:

1. Software engineering / coding
2. Research and writing
3. Other

A task should be classified as category 1 or 2 if the actual work primarily involves these skills, such that with sufficient knowledge one could solve the task by just using these skills.

Examples of category 1:
- Front-end development
- Game development
- Website creation

Examples of category 2:
- Reading PDFs and writing a report
- Searching for information online and writing a report
- Writing a blog post about a historical event

Examples of category 3:
- Performing research, running simulations, and writing a report
- Making an as-built drawing of a building
- Creating an educational video
- QA testing for a video game and writing a bug report (involves playing the game)
\end{promptblock}

For HCAST, we manually classify the task distribution shown in Table 1 of the HCAST paper \cite{rein2025hcast}. For an estimate of the Upwork distribution, we apply the above prompt to the category names in the Upwork taxonomy. This provides a distribution over the different types of work performed on Upwork. Note: This is not a distribution at the job-level, which is more skewed toward software tasks.

\subsection{Data Collection Details}\label{app:data_collection}

\begin{enumerate}
    \item For projects sourced from freelancers, we only included projects where freelancers explicitly verified that they had the rights to sell us the work.
    \item In cases where the work contains PII or copyrighted content (e.g., logos or company names), we anonymized the project by redacting information. In some cases, redacted information was replaced with synthetic details (e.g., fake company names or logos).
    \item For long-tail project collection, we either purchased the work or received permission from the original author of the work to link to it in our study.
\end{enumerate}

\subsection{Project cost and completion time.}
\label{app:time_cost_distribution}

\paragraph{Collecting cost and completion time.} For the vast majority of projects, the human professionals who created the human deliverable provided the cost and completion time for the project. These metrics were operationalized as follows:

\begin{itemize} \item \textbf{Cost:} The amount of money in USD earned by the freelancer for completing the project, or a fair price estimated by the professional for recreating the work from scratch. Human professionals self-reported these values. Since these often represent the actual amount of money paid by a client, they provide an accurate measure of the cost of the project. \item \textbf{Completion time:} The amount of time in hours that it took human professionals to complete the projects. These values were also self-reported to ensure economic accuracy. \end{itemize}

In some cases, human professionals communicated a range of times or costs; in these instances, we took the midpoint value. Costs are available for $95\%$ of projects. Completion times are available for $84\%$ of projects. $5\%$ of projects have neither cost nor completion time data, but were kept in the dataset due to being high-quality. For experiments or metrics using this data, we drop projects for which the required values are not available.

\paragraph{Distribution over project cost and completion time.}

In \Cref{fig:completion-time-cost}, we show the distributions over project cost and completion time. Both variables are roughly log-normal distributed, with project cost and completion time reaching up to $\$22,\!500$ and $450$ hours. Individual numbers are often rounded by freelancers who self-report the data, and fixed price projects tend to cluster at whole-number values, explaining peaks in the data.

\begin{figure}[t]
\centering
\includegraphics[width=0.7\linewidth]{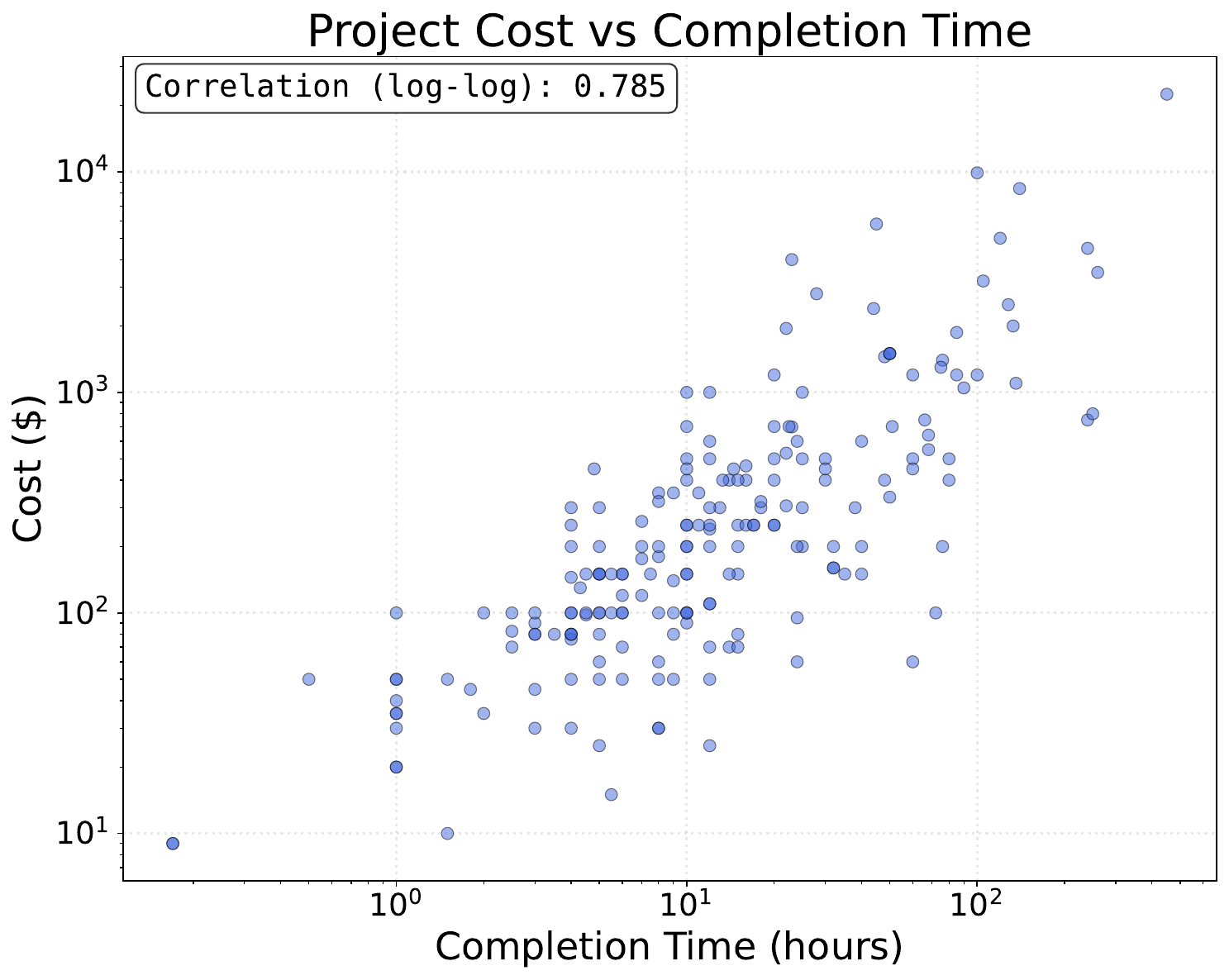}
\caption{Project cost and completion time are highly correlated on a log-log scale.}
\label{fig:completion-time-cost-scatter}
\end{figure}

In \Cref{fig:completion-time-cost-scatter}, we plot these variables against each other on a log-log scale for projects where both values are available. We observe a Pearson correlation of $0.785$.

\subsection{AI Deliverable Examples}
\label{app:ai-examples}

\begin{figure}[h]
\centering
\includegraphics[width=1.0\linewidth]{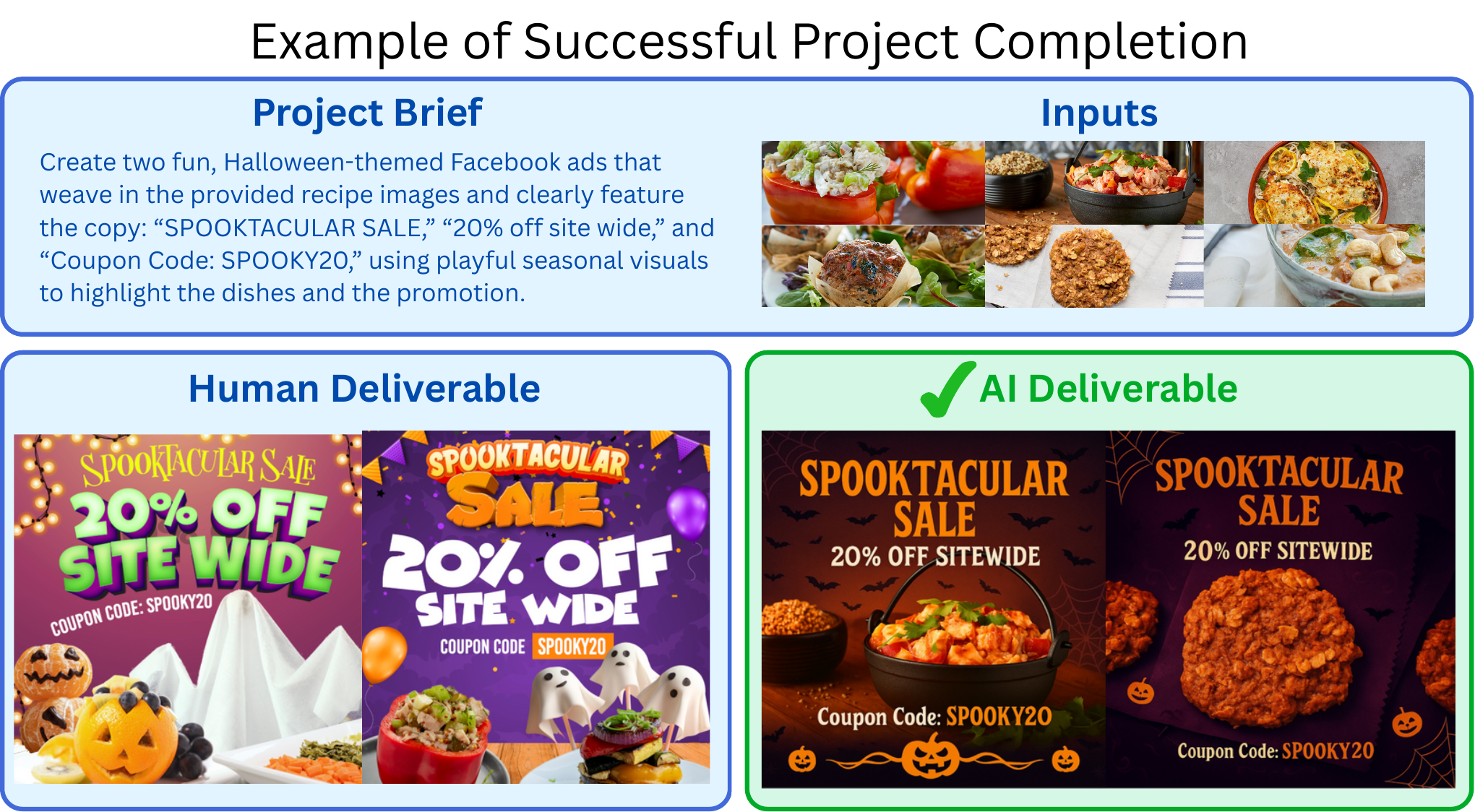}
\caption{AI agents leverage image generation tools to solve some marketing projects in RLI. Here we show a successful project completion from Manus.}
\label{fig:ai-deliverable-examples-1}
\end{figure}

\begin{figure}[h]
\centering
\includegraphics[width=1.0\linewidth]{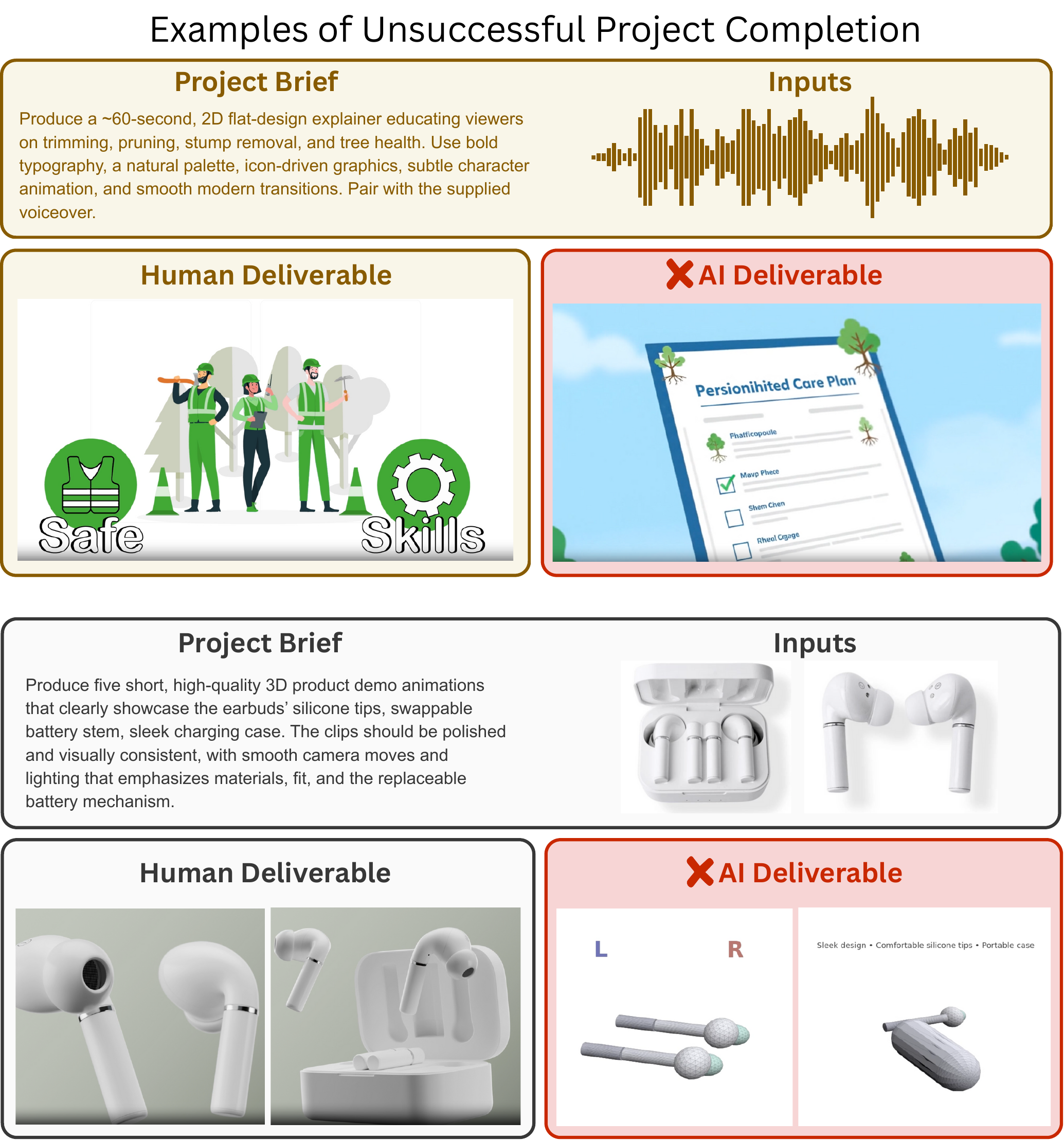}
\caption{Agents fail to successfully complete the vast majority of RLI projects. Here we show failed projects for Gemini 2.5 Pro (top) and GPT-5 (bottom).}
\label{fig:ai-deliverable-examples-2}
\end{figure}

\FloatBarrier
\subsection{Detailed Project Examples}

\begin{figure}[h]
\vspace{-10pt}
\centering
\includegraphics[width=0.95\linewidth]{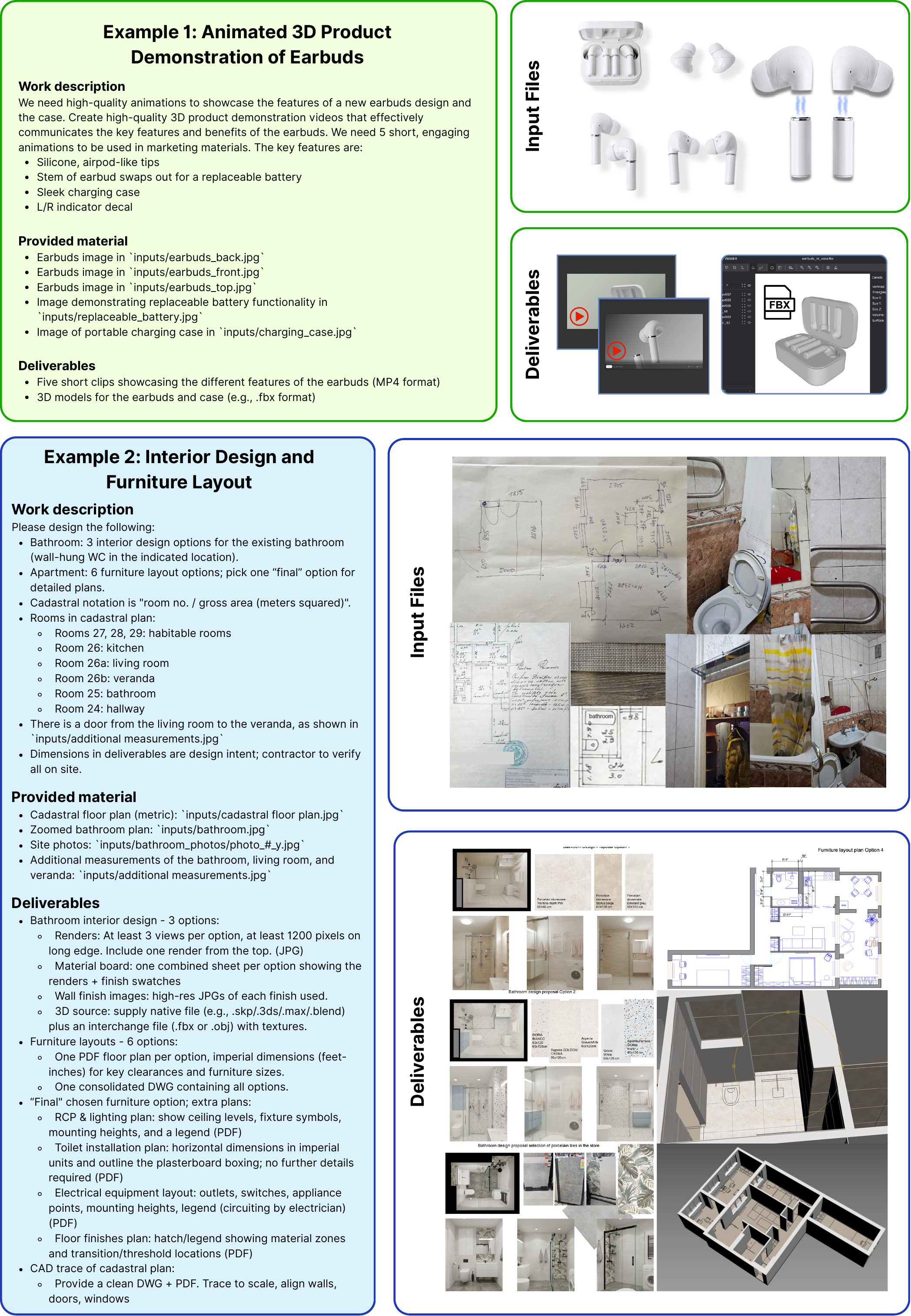}
\caption{Detailed project examples with extended briefs.}
\label{fig:detailed-project-examples-1}
\end{figure}

\begin{figure}[h]
\centering
\includegraphics[width=0.95\linewidth]{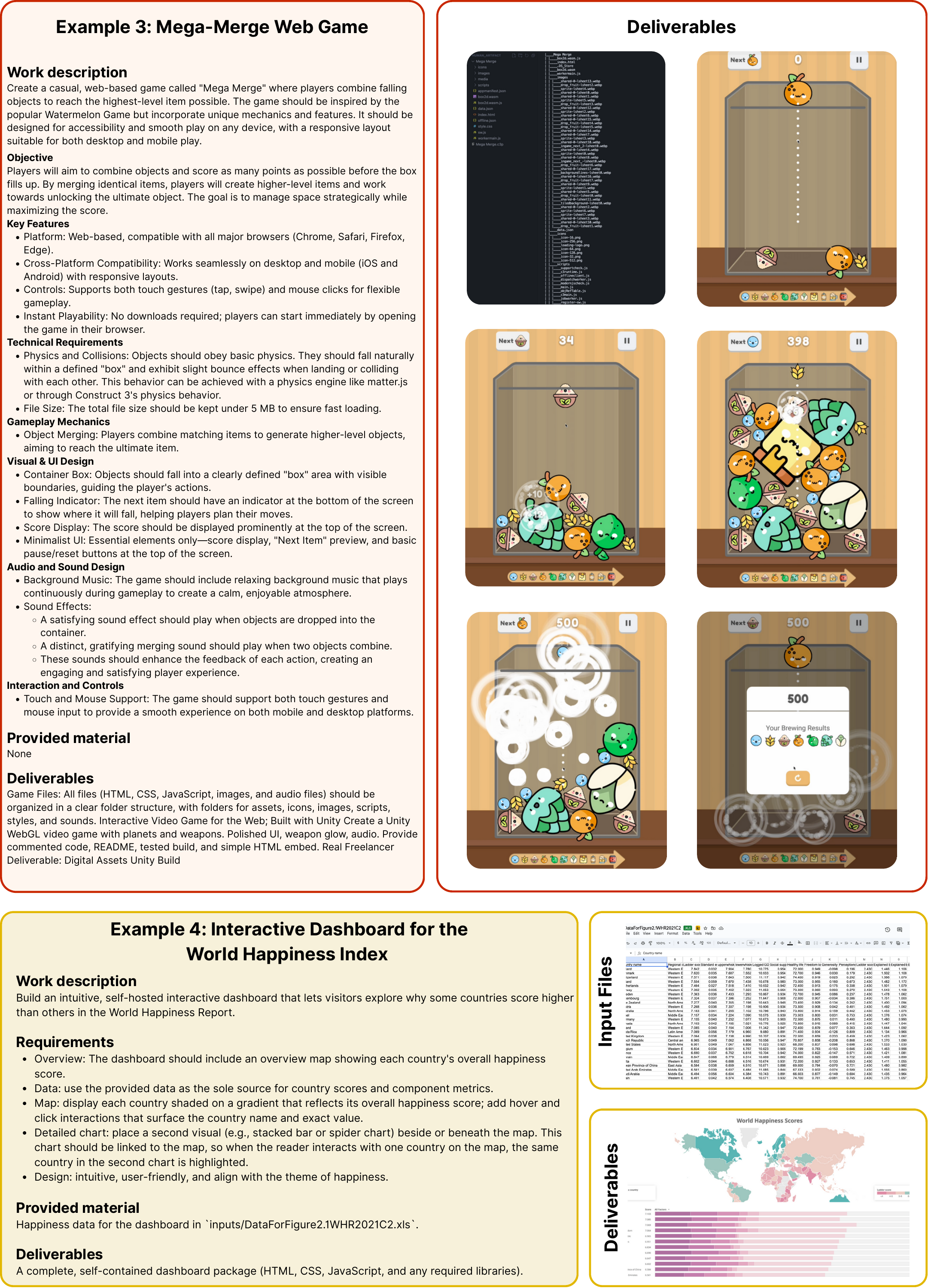}
\caption{Detailed project examples with extended briefs.}
\label{fig:detailed-project-examples-2}
\end{figure}

\end{document}